\documentclass[sigconf, nonacm]{acmart}
\AtBeginDocument{%
  }

\setcopyright{acmlicensed}
\copyrightyear{2025}
\acmYear{2025}
\acmDOI{XXXXXXX.XXXXXXX}
\acmConference[ACM MMAsia 2025]{ACM Multimedia Asia 2025}{December 09--12,
  2025}{Kuala Lumpur, Malaysia}
\acmISBN{978-1-4503-XXXX-X/2018/06}

\acmSubmissionID{27}

\usepackage{caption}  
\usepackage{subcaption}
\usepackage{algorithm}
\usepackage{algorithmic}
\usepackage{multirow}
\usepackage{booktabs, array}
\usepackage{textcomp}
\usepackage{bbding}
\begin{document}

\title{Mixture of Group Experts for Learning Invariant Representations}

\author{Lei Kang}
\affiliation{%
	\institution{Beijing Normal University}
	\city{Beijing 100875}
	\country{China}}
\email{leikang@mail.bnu.edu.cn}

\author{Jia Li}
\authornote{Also with the Engineering Research Center of Intelligent Technology and Educational Application, Ministry of Education, Beijing 100816, China}
\affiliation{%
	\institution{Beijing Normal University}
	\city{Beijing 100875}
	\country{China}}
\email{jiali@bnu.edu.cn}

\author{Mi Tian}
\affiliation{%
	\institution{TAL Education Group}
	\city{Beijing 102206}
	\country{China}}
\email{tianmi@tal.com}

\author{Hua Huang}
\authornotemark[1]
\affiliation{%
	\institution{Beijing Normal University}	
	\city{Beijing 100875}
	\country{China}}
\email{huahuang@bnu.edu.cn}
%
%
%\affiliation{%
%	\institution{School of Artificial Intelligence, Beijing Normal University}
%	\city{Beijing 100875}
%	\country{China}}
%
%\affiliation{%
%	\institution{Engineering Research Center of Intelligent Technology and Educational Application}
%	\city{Beijing 100816}
%	\country{China}}
%
%\affiliation{%
%	\institution{TAL Education Group}
%	\city{Beijing 102206}
%	\country{China}}

\renewcommand{\shortauthors}{Kang et al.}

%%
%% Abstract + CCS Concepts + Keywords
\begin{abstract}
	Sparsely activated Mixture-of-Experts (MoE) models effectively increase the number of parameters while maintaining consistent computational costs per token. However, vanilla MoE models often suffer from limited diversity and specialization among experts, constraining their performance and scalability, especially as the number of experts increases.  In this paper, we present a novel perspective on vanilla MoE with top-$k$ routing inspired by sparse representation. This allows us to bridge established theoretical insights from sparse representation into MoE models. Building on this foundation, we propose a group sparse regularization approach for the input of top-$k$ routing, termed Mixture of Group Experts (MoGE). MoGE indirectly regularizes experts by imposing structural constraints on the routing inputs, while preserving the original MoE architecture. Furthermore, we organize the routing input into a 2D topographic map, spatially grouping neighboring elements. This structure enables MoGE to capture representations invariant to minor transformations, thereby significantly enhancing expert diversity and specialization. Comprehensive evaluations across various Transformer models for image classification and language modeling tasks demonstrate that MoGE substantially outperforms its MoE counterpart, with minimal additional memory and computation overhead.  Our approach provides a simple yet effective solution to scale the number of experts and reduce redundancy among them. The source code is included in the supplementary material and will be publicly released.
\end{abstract}





\maketitle

% Introduction
\section{Introduction}
\label{sec:introduction}

\begin{figure}[t]
	\centering
	\begin{subfigure}{1\linewidth}
		\includegraphics[width=1\linewidth]{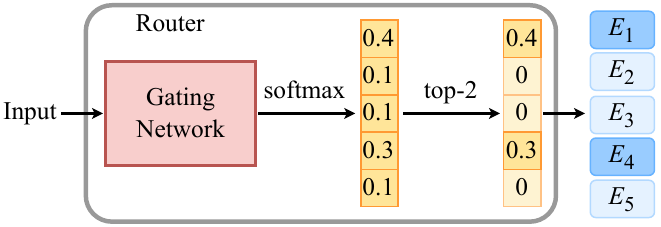}
		\caption{Vanilla top-2 routing MoE layer}
		\label{subfig:MoE}
	\end{subfigure}
	\vfill
	\begin{subfigure}{1\linewidth}
		\includegraphics[width=1\linewidth]{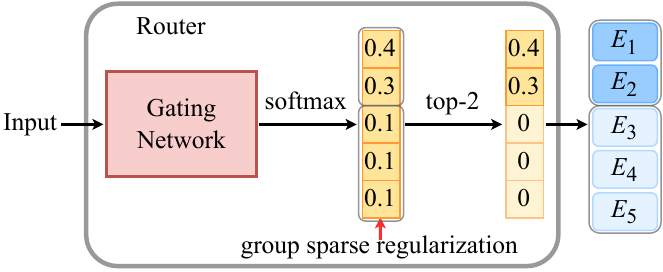}
		\caption{Our proposed top-2 routing MoGE layer}
		\label{subfig:MoGE}
	\end{subfigure}
	\caption{A high-level comparison between the vanilla MoE layer and the proposed MoGE layer. (a) illustrates the vanilla MoE layer with the top-2 routing, while (b) demonstrates the application of group sparse regularization in MoGE. It is important to note that the using of group sparse regularization does not alter the MoE architecture. In the vanilla MoE, each element in the input of top-2 routing is independent and considered part of a single group. In contrast, MoGE introduces a grouping mechanism that partitions the input of top-2 routing into two groups: the first two elements in one group and the remaining three elements in the other. This grouping effect naturally extends to the subsequent experts $\{E_1,\cdots,E_5\}$, resulting in their corresponding organization into same groups.}
	\label{fig:MoEVSMoGE}
\end{figure}

% ********************* MoE 很重要 *********************
Transformer models~\cite{vaswani2017attention} have consistently demonstrated performance improvements as their parameter counts increase~\cite{kaplan2020scaling}. However, the growth in model size comes with significant computational costs, making further scaling increasingly challenging. Sparsely activated Mixture-of-Experts (MoE)~\cite{shazeer2017outrageously} offers a promising solution by utilizing a sparse architecture that activates only a subset of parameters during both training and inference. As shown in Fig.~\ref{fig:MoEVSMoGE} (a), an MoE layer comprises multiple experts, activating only a few experts for each input token. A gating network is trained to route each token to the most suitable experts. This conditional computation enables MoE models to significantly increase their capacity without a rise in computation overhead. By integrating MoE layers into Transformer models, researchers have successfully scaled large foundation models to  impressive sizes, achieving outstanding performance in tasks such as image classification\cite{riquelme2021scaling, hwang2023tutel} and language modeling~\cite{teo2024momentumsmoe, csordas2024switchhead}.

% ********************* MoE experts 易陷入相同的分布，diversity研究很多 *********************
Despite their advantages, the vanilla MoE models suffer from insufficient diversity and specialization among experts~\cite{sun2024mixture,wang2024hmoe,feng2025omoe}. In each MoE layer, the gating network determines token-to-expert assignments and should route similar tokens to the same experts to promote diversity and specialization. While load balancing losses~\cite{shazeer2017outrageously,lepikhingshard2021} encourage even token distribution, they do not ensure that similar tokens activate similar experts. As the number of experts grows, promoting diversity and specialization remains a  challenge for MoE models.

\noindent\textbf{Contributions:} In this paper, we reinterpret the combination of the most popular \mbox{top-$k$} routing and experts as a form of sparse representation. Building on this perspective, we propose a novel regularization technique that is applied to the routing inputs, rather than directly constraining the experts themselves, to promote greater diversity and specialization. Our approach does not require modifications to the MoE architecture and can be directly applied to any MoE model. The main contributions are highlighted below.
 
Firstly, we propose investigating MoE employing \mbox{top-$k$} routing from the perspective of  sparse representation. Sparse representation aims to find the sparse code that effectively approximates the input when multiplied with the dictionary~\cite{donoho2003optimally,tropp2004greed}. For the MoE with \mbox{top-$k$} routing, the routing output can be viewed as the sparse code, while the output of each  expert corresponds to a column in the dictionary. This perspective enables the application of the rich theoretical results of sparse representation to MoE.

Secondly, we propose introducing group sparse regularization to the inputs of \mbox{top-$k$} routing to indirectly impose structural constraints on experts.  Group sparsity assumes that the input elements of \mbox{top-$k$} routing are organized into predefined groups, where all elements within a group tend to be either all zeros or all nonzeros~\cite{yuan2006model,eldar2010block}. These groups are typically formed based on an arbitrary partition, informed by prior knowledge specific to the task. As illustrated in Fig.~\ref{fig:MoEVSMoGE} (b), we refer to the application of group sparse regularization in MoE as the Mixture of Group Experts (MoGE). Inspired by previous works~\cite{hyvarinen2001two,hyvarinen2007complex,kavukcuoglu2009learning}, we further organize the routing input into a 2D matrix, forming a topographic map where neighboring elements are grouped together to activate similar experts. This topographically organized input enables MoGE to extract locally invariant representations, encouraging diversity and specialization among experts.

Finally, we evaluate our MoGE across various image classification and language modeling tasks. Consistently, our MoGE demonstrates superior performance compared to its MoE counterpart, with only a negligible increase in memory consumption and computational cost. Additionally, our MoGE effectively promotes diversity and specialization among experts through an easily implementable approach, enabling the training of a greater number of experts and larger $k$ values in \mbox{top-$k$} routing.

% Methodology
\section{Methodology}
\label{sec:methodology}

\subsection{Vanilla MoE}
\label{sec:VanillaMoE}

To recap the core concepts behind our MoGE, we focus specifically on the sparsely activated MoE layer and the \mbox{top-$k$} routing. Existing MoE models are typically derived from dense Transformer models by replacing certain feedforward layers with MoE layers~\cite{riquelme2021scaling, fedus2022switch, jiang2024mixtral}. As illustrated in Fig.~\ref{fig:MoEVSMoGE} (a), an MoE layer generally consists of a set of experts $\{E_1(\cdot), \cdots, E_n(\cdot)\}$ and a router that directs input tokens to the most relevant experts. Here, $n$ denotes the total number of experts, and $E_i(\cdot) \in \mathbb{R}^m$ for $i=1,\cdots,n$.  

The router leverages the \mbox{top-$k$} routing mechanism to reduce computation overhead by selectively activating only the \mbox{top-$k$} experts deemed most relevant to each input token. It typically includes a gating network $G(\cdot)$, a softmax function, and a \mbox{top-$k$} function. For any input token $x$, the routing decision is computed as follows:
\begin{align}\label{eq:gatingnetwork}
	z = \mathrm{softmax}(G(x)),
\end{align}
where $z \in \mathbb{R}^n$. Subsequently, sparsely activated weights are computed using:
\begin{align}\label{eq:forcesparsity}
	w = \mathrm{TopK}(z, k),
\end{align}
where the \mbox{top-$k$} function, $\mathrm{TopK}(\cdot,\cdot)$, is defined as:
\begin{align}
	\mathrm{TopK}(v, k) =& \begin{cases}v_i, &\text{if } v_i \ \text{is among the top-$k$ elements of } v;\\ 0, &\text{otherwise}.\end{cases}.
\end{align}
The final output $y \in \mathbb{R}^m $ of the MoE layer is computed as a linear combination of the outputs from the selected experts weighted by the sparsely activated weights:
\begin{equation}\label{eq:weighted_sum}
	y = \sum_{i \in \mathcal{I}}  w_i E_i(x),
\end{equation}
where $\mathcal{I}$ denotes the set of \mbox{top-$k$} indices derived from $z$.

%For simplicity, we omit considerations of noise, load balancing loss, and any sequence dependencies between the softmax and the \mbox{top-$k$} functions. In our experiments, we follow the default settings for each MoE model outlined in its original paper.

\subsection{Sparse Representation for MoE}

Given an overcomplete dictionary $\mathcal{D} \in \mathbb{R}^{m \times n}~(m < n)$, sparse representation~\cite{donoho2003optimally,tropp2004greed,wright2010sparse} expresses an input signal $y \in \mathbb{R}^m$ as the product of the dictionary $\mathcal{D}$ and a sparse code $w \in \mathbb{R}^n$. Namely, $y$ is represented as a linear combination of the columns of the dictionary $\mathcal{D}$, weighted by elements in the sparse code $w$. When the number of nonzero elements in $w$ is constrained to be no greater than $k$, the sparse representation problem can be formulated as:
\begin{equation}
	\min_{w}\frac{1}{2}\|y-\mathcal{D}w\|_2^2,\quad\text{s.t.}~ \|w\|_0 \leq k,
	\label{eq:sparsecoding}
\end{equation}
where $\Vert w \Vert_0$ denotes the count of nonzero
elements in $w$.

By treating $E_i(x)$ as the $i$-th column of the dictionary $\mathcal{D}$, Eqs.~\eqref{eq:forcesparsity} and~\eqref{eq:weighted_sum} can be interpreted as the solution to problem~\eqref{eq:sparsecoding}. Consequently, the MoE layer utilizing \mbox{top-$k$} routing can be viewed as addressing a sparse representation problem. Sparse representation, having been developed over the past two decades, offers a wealth of validated theoretical foundationsthat can be effectively leveraged in the design of MoE models.

In this paper, we adopt the simplest approach by promoting the diversity and specialization of experts through the application of regularization to the sparse code $w$. However, due to the limited number of experts, the $k$ value is usually small, with $k=1$  and $k=2$  being the most commonly used. The high sparsity of nonzero elements in $w$  diminishes the effectiveness of regularization. To address this, we propose applying regularization to $z$ in~\eqref{eq:gatingnetwork}, which serves as the input to the top-$k$ function.

\begin{figure*}[t]
	\centering
	\includegraphics[width=1.0\linewidth]{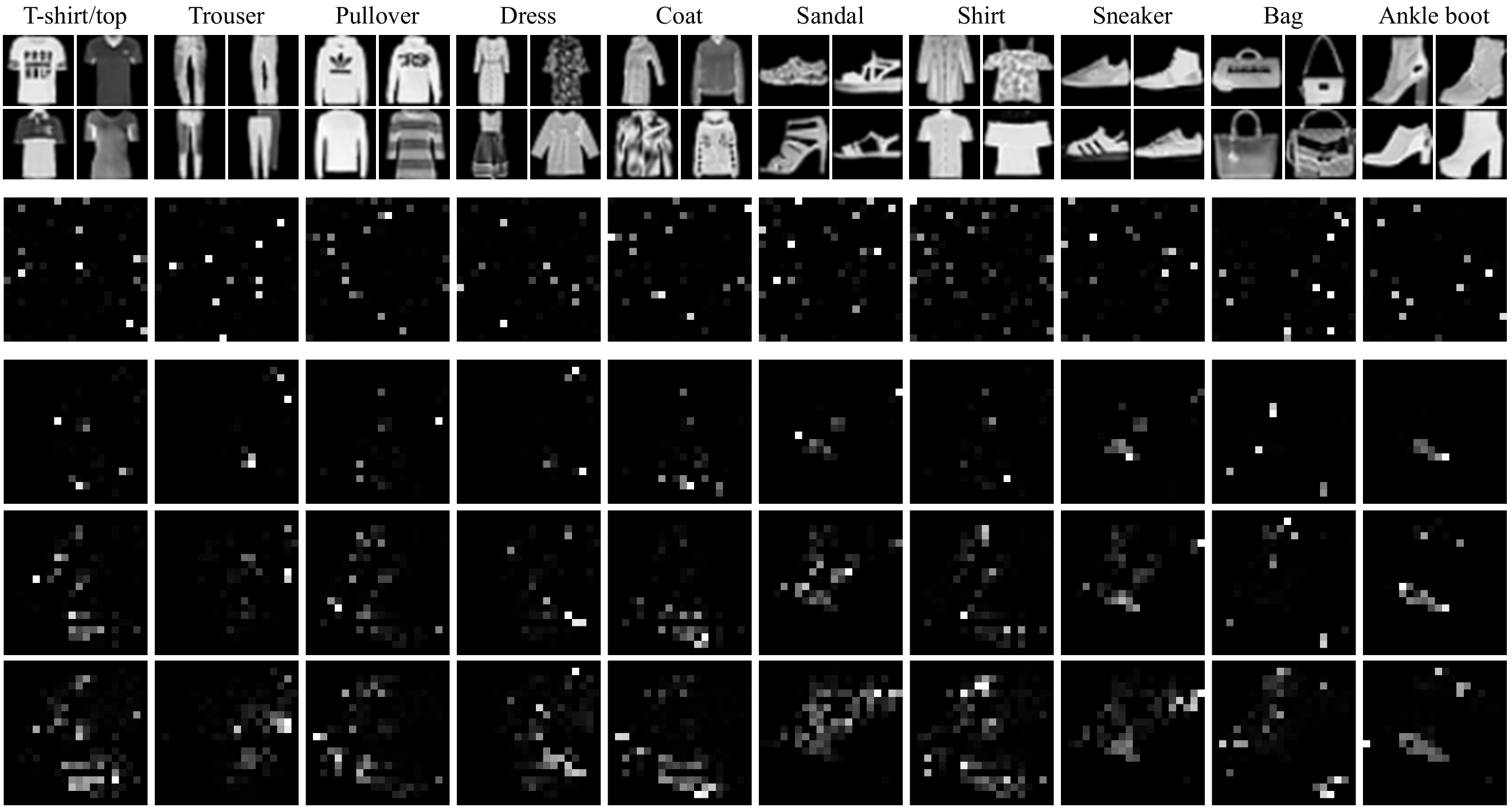}
	\caption{The average inputs of top-1 routing in 2D (denoted as \( Z \) in Algorithm~\ref{alg:MoGE}) of MoE and MoGE for each category in the Fashion-MNIST test set. The first row corresponds to the ten categories in the Fashion-MNIST dataset. The second row shows the average of \( Z \) for MoE across each category at the 150th epoch. The subsequent three rows display the average of $Z$ for MoGE across each category at the 50th, 100th, and 150th epochs, respectively. 	We can see that the average elements of \( Z \) in MoGE demonstrate a clear aggregation effect throughout the training process. In contrast, such an aggregation effect is absent in MoE. }   
	\label{fig:DistributionOfMoEandMoGE}
\end{figure*}

\begin{algorithm}
	\caption{Computing our group sparse regularization.}
	\label{alg:MoGE}
	\begin{algorithmic}[1] %这个1 表示每一行都显示数字
		\REQUIRE input of \mbox{top-$k$} routing $z \in \mathbb{R}^n$, size $h$ and standard deviation $\sigma$ of Gaussian lowpass filter.\\
		\ENSURE  group sparse regularization $R(z) \in \mathbb{R}$.
		\STATE Reshape $z \in \mathbb{R}^n$ into an approximate square matrix $Z \in \mathbb{R}^{r \times c}$ such that $n=r\times c$ and $r$ is close to $c$		
		\STATE Compute $\tilde{Z} = Z\odot Z$
		\STATE Generate Gaussian filter $F_{\sigma} \in \mathbb{R}^{h \times h}$
		\STATE Compute the 2D convolution $\tilde{P}=F_{\sigma} * \tilde{Z}$
		\STATE Compute $P = \sqrt{\tilde{P}}$
		\STATE Compute group sparse regularization $R(z)=\sum_i\sum_jP_{ij}$
		%\RETURN $R(z)$;
	\end{algorithmic}
\end{algorithm}

\subsection{Mixture of Group Experts}
 
Adding structural constraints to sparse codes has been demonstrated as an effective strategy. In the original problem presented in~\eqref{eq:sparsecoding}, any column of the dictionary can be selected freely. However, in the structured sparse model~\cite{yuan2006model,eldar2010block}, instead of selecting individual columns, groups of columns are selected collectively. These groups can overlap and vary in size, with the goal of representing $y$ such that the code $w$ remains sparse while adhering to the imposed structure. Typically, these groups are defined based on prior knowledge relevant to the task. We refer to the incorporation of group sparse regularization in MoE as the Mixture of Group Experts (MoGE) (see Fig.~\ref{fig:MoEVSMoGE} (b)).

Designing a structural constraint for sparse codes that performs effectively across diverse tasks is a significant challenge. A widely adopted general structural constraint is local invariance, which ensures that the output remains consistent when the input undergoes minor transformations. For instance, in image classification tasks, small perturbations of the input images such as translations, rotations, scalings, or degradations should not drastically affect the extracted features. In the context of MoGE, we strive for the corresponding representation $z$ to exhibit invariance under small changes to the input token $x$. This property ensures that the selection of experts remains consistent, thereby mapping similar tokens to similar experts. Since these experts process analogous tokens as input, they can  enhance diversity and specialization. 

\begin{table*}[t]
	\centering  
	\caption{Comparison of the average image Euclidean distances in $Z$ between the original images from the Fashion-MNIST test set and their corresponding variations under different transformations. $\downarrow$ indicates that smaller values are better.}  
	\label{tab:PytorchTransforms}
	\begin{tabular*}{\textwidth}{@{\extracolsep{\fill}}c ccc|ccc|ccc|ccc}  
		\toprule  
		\multirow{2}{*}{Type} & \multicolumn{3}{c}{Rotation $\downarrow$} & \multicolumn{3}{c}{Scaling $\downarrow$} & \multicolumn{3}{c}{Translation $\downarrow$} & \multicolumn{3}{c}{Shear $\downarrow$} \\ 
		\cmidrule(lr){2-13}
		~      & 5     & 10     & 15            & 0.5      & 0.8     & 1.1              & (0, 0.1)   & (0.1, 0)   & (0.1, 0.1) & 5     & 10     & 15    \\ 
		
		\midrule  
		MoE   & 0.0778 & 0.1301 & 0.1670        & 0.2359 & 0.2268 & 0.1466         & 0.1181 & 0.1218 & 0.1759         & 0.0345 & 0.0671 & 0.0980   \\   
		MoGE  & \textbf{0.0473} & \textbf{0.0811} & \textbf{0.1067}        & \textbf{0.1948} & \textbf{0.1680} & \textbf{0.1147}         & \textbf{0.0786} & \textbf{0.0809} & \textbf{0.1206}         & \textbf{0.0209} & \textbf{0.0396} & \textbf{0.0576}   \\ 
		\bottomrule  
	\end{tabular*}  
\end{table*}

Building upon the foundational studies of the visual cortex~\cite{hyvarinen2001two,hyvarinen2007complex,kavukcuoglu2009learning}, we propose a topographically organized regularization for $z$ aimed at facilitating the learning of invariant representations. Specifically, our approach employs group sparse regularization, structuring $z$ into a 2D matrix to form a topographic map. This arrangement encourages neighboring elements to activate similar experts, thereby preserving local invariance. Moreover, this design enhances both the consistency and interpretability of expert selection, offering a robust framework for structured representation learning.

We first introduce the computation process of our group sparse regularization for \(z\) and then provide an analysis of the underlying reasoning. Given a vector \(z \in \mathbb{R}^n\), we begin by reshaping it into an approximately square matrix \(Z \in \mathbb{R}^{r \times c}\), where \(n = r \times c\) and \(r\) is chosen to be as close to \(c\) as possible. Next, we compute the elementwise square of \(Z\) to obtain \(\tilde{Z}\), i.e., \(\tilde{Z} = Z \odot Z\), where \(\odot\) denotes the elementwise product. Subsequently, we generate a rotationally symmetric Gaussian lowpass filter \(F_{\sigma}\) of size \(h \times h\) with a standard deviation \(\sigma\). We then convolve \(\tilde{Z}\) with the Gaussian filter \(F_{\sigma}\) to produce \(\tilde{P}\), i.e., $\tilde{P}=F_{\sigma}*\tilde{Z}$, where $*$ denotes the 2D convolution. Finally, we compute the square root of each element in \(\tilde{P}\) to obtain \(P\). The group sparse regularization \(R(z)\) is defined as the sum of all elements in \(P\). We summarize the entire computation process of our group sparse regularization in Algorithm~\ref{alg:MoGE}.

To illustrate the mechanism of our group sparse regularization, we first introduce the general form of \( l_{p,q} \) group sparse regularization~\cite{hu2017group}. Consider the group structure of \( z \),  represented as \( \{z_{\mathcal{G}_1}, \cdots, z_{\mathcal{G}_s}\} \). The \( l_{p,q} \) group sparse regularization of \( z \) is defined as follows:  
\begin{equation}\label{eq:lpq}
	\|z\|_{p,q} = \left(\sum_{i=1}^s \|z_{\mathcal{G}_i}\|_p^q\right)^{1/q},
\end{equation}  
where $s$ denotes the number of groups, \( p \geq 1 \), \( 0 \leq q \leq 1 \), and \( \|\cdot\|_p \) denotes the \( l_p \)-norm for \( p > 0 \), defined as \( \|z\|_p = \left(\sum_{i} |z_i|^p\right)^{1/p} \). The \( l_{p,q} \) group sparse regularization encourages all elements within each group to be either entirely zero or entirely nonzero. A widely used special case is the \( l_{2,1} \) regularization, defined as:  
\begin{equation}\label{eq:lpq}
	\|z\|_{2,1} = \sum_{i=1}^s \|z_{\mathcal{G}_i}\|_2.
\end{equation} 
In Algorithm~\ref{alg:MoGE}, when we employ an average filter  of size \( h \times h \) instead of a Gaussian filter, it is clear that our group sparse regularization corresponds to \( l_{2,1} \) regularization, where all elements within each local neighborhood of size \( h \times h \) are vectorized and assigned to a group and spatially adjacent groups are overlapping. This design enables our group sparse regularization to cluster neighboring elements in \( Z \), thereby activating similar experts.  We simply set \( p = 2 \) and \( q = 1 \) in our current approach. Other configurations satisfying \( p \geq 1 \) and \( 0 \leq q \leq 1 \) are also applicable. A thorough investigation of the effects of varying \( p \) and \( q \) is deferred to future work.

We use a Gaussian filter instead of an average filter for the following two key reasons. First, unlike the average filter, the Gaussian filter preserves the 2D structural information within the group by incorporating the distance from the center into its calculations. Second, the Gaussian filter provides the flexibility to adjust the standard deviation $\sigma$. A larger $\sigma$ facilitates sufficient competition within the group, while a smaller $\sigma$ encourages aggregation. In practice, the standard deviation $\sigma$ is not fixed and can follow a dynamic scheduling strategy, enabling it to adapt throughout the training process. We aim to promote sufficient competition among group elements during the early stages of training, while encouraging them to become more concentrated as training progresses. To achieve this, we employ the following pow function scheduling:
\begin{equation}\label{eq:DynamicSigma}
\sigma_t = \sigma_{\text{0}} - (\sigma_{0} - \sigma_{\min})~\left(\frac{t}{T}\right)^{\gamma},
\end{equation}
where $\sigma_{0}>0$ and $\sigma_{\min}>0$ are constants satisfying $\sigma_{0} \geq \sigma_{\min}$, $\gamma \geq 0$, $t$ represents the current iteration number, $\gamma$ is the exponent, and $T$ is the total number of iterations. The curves of $\sigma$ with varying $\gamma$ over iterations are illustrated in  Appendix~A. Finding the optimal scheduler for $\sigma$ is a challenging task. %This complexity arises not only from the intricate interplay between learning rate, batch size, number of experts $n$, the $k$ value, and other hyperparameters but also from the prohibitively high cost of conducting a hyperparameter search for MoE models with massive parameters~\cite{bergsma2025straight}. 
We will explore optimal scheduler for $\sigma$ in the future.

We incorporate the proposed group sparse regularization \( R(z) \) into the vanilla MoE with its original loss function \( \mathcal{L} \), resulting in the following loss function for MoGE:
\begin{equation}
	\mathcal{L} + \lambda~R(z), 
	\label{eq:MoGE_finalloss}
\end{equation}
where \( \lambda > 0 \) balances the original loss function and the regularization, and is set as described in Appendix~B. Our group sparse regularization requires computing a 2D convolution with a Gaussian lowpass filter (see Algorithm~\ref{alg:MoGE}). In practice, we empirically use valid convolution to avoid padding operations on $\tilde{Z}$. Additionally, we need to predefine the filter size \(h\) and the standard deviation \(\sigma\). We conduct ablation experiments in subection~\ref{ablation:StaticVSDynamicSigma}.

\subsection{Investigating Invariant Representations}

We explore how MoGE learns invariant representations by employing a simplified model with a single MoE layer for image classification on the Fashion-MNIST dataset~\cite{xiao2017fashion}. The model consists of 400 homogeneous experts with top-1 routing, where each image is treated as a token. Each expert is a two-layer feedforward network, featuring 784 neurons in both the input and output layers, and 64 neurons in the hidden layer. The output of the MoE layer is fed directly into a single-layer feedforward classifier for categorization.

We train MoE and MoGE models from scratch for 150 epochs, setting $h=3$, \(\sigma = 2\), and $\lambda =$ \mbox{4e-3} for MoGE. The top-1 test accuracies achieved by MoE and MoGE are \(41.70\%\) and \(44.74\%\), respectively. Due to the large number of experts and the relatively small size of the training set, these test accuracies are not particularly high. However, our primary focus here is on the relative performance of the two models. Fig.~\ref{fig:DistributionOfMoEandMoGE} illustrates the average \( Z \) values in Algorithm~\ref{alg:MoGE} for both MoE and MoGE across each category in the Fashion-MNIST test set. Notably, the average elements of \( Z \) in MoGE exhibit a clear aggregation effect throughout the training process. In contrast, this aggregation effect is not observed in MoE. This indicates that the experts activated for each category in MoGE exhibit spatial similarity, which helps enhance expert diversity and specialization.

To validate that MoGE effectively learns invariant representations, we apply various transformations to the images in the Fashion-MNIST test set and evaluate the image Euclidean distances (IMED)~\cite{wang2005euclidean} in $Z$ between the original and the transformed images. IMED accounts for spatial correlations among pixels, avoiding the perceptual pitfalls of the standard Euclidean distance, and is widely used for 2D image comparison. As shown in Table~\ref{tab:PytorchTransforms}, we use the default function \texttt{RandomAffine} in PyTorch to implement four transformations: rotation, scaling, translation, and shear. The results reported in Table~\ref{tab:PytorchTransforms} demonstrate that following these transformations, the average IMED values in $Z$ for MoGE are consistently smaller than those for MoE. This highlights the significant advantage of MoGE in learning more invariant representations.

% Experiments
\section{Experiments}
\label{sec:experiments}

In this section, we evaluate the superiority of MoGE on image classification and language modeling tasks, followed by ablation studies.% Subsequently, we test the scalability and efficiency of MoGE, followed by ablation studies. % on the Gaussian lowpass filter. 

%-------------------------------------------------------------------------

\begin{table}[t]
	\caption{Comparison of MoE and our MoGE for pretraining and fine-tuning on image classification.}
	\label{tab:ResultsOfImageClassification}
	\centering
	\begin{tabular*}{\columnwidth}{@{\extracolsep{\fill}}c p{2.25cm} cccc@{}}
		\toprule
		Dataset & \centering Model & $n$ & $k$ & Acc@1 & Acc@5 \\
		\midrule
		% ---------------- CIFAR-100 ----------------
		\multirow{4}{*}{CIFAR-100}
		%        \multirow{4}{*}{C100}
		& \raggedright V-MoE-S								& 32 & 1 & 74.03          & 91.49          \\
		& \raggedright V-MoGE-S                          	& 32 & 1 & \textbf{74.67} & \textbf{91.55} \\
		\cmidrule(lr){2-6}
		& \raggedright SwinV2-MoE-T							& 32 & 1 & 77.35          & 92.41          \\
		& \raggedright SwinV2-MoGE-T                     	& 32 & 1 & \textbf{77.75} & \textbf{92.61} \\
		\midrule
		
		% ---------------- Tiny-ImageNet ----------------
		\multirow{4}{*}{Tiny-ImageNet}
		%        \multirow{4}{*}{TINET}	
		& \raggedright V-MoE-S								& 32 & 1 & 55.38          & 76.66          \\
		& \raggedright V-MoGE-S                          	& 32 & 1 & \textbf{56.15} & \textbf{77.24} \\
		\cmidrule(lr){2-6}
		& \raggedright SwinV2-MoE-T							& 32 & 1 & 59.01          & 78.97          \\
		& \raggedright SwinV2-MoGE-T                     	& 32 & 1 & \textbf{60.19} & \textbf{79.50} \\
		\midrule
		
		% ---------------- ImageNet-1K (scratch) ----------------
		\multirow{4}{*}{ImageNet-1K}
		%        \multirow{4}{*}{INET1K}
		& \raggedright V-MoE-S								& 32 & 1 & 76.95 		  & 92.95		   \\
		& \raggedright V-MoGE-S                             & 32 & 1 & \textbf{77.24} & \textbf{93.18} \\
		\cmidrule(lr){2-6}
		& \raggedright SwinV2-MoE-T							& 32 & 1 & 79.14 & 94.00 \\
		& \raggedright SwinV2-MoGE-T                        & 32 & 1 & \textbf{79.97} & \textbf{94.94}  \\
		\midrule
		
		% ---------------- ImageNet-1K (fine-tune) ----------------
		%        \multirow{4}{*}{\parbox[c]{2.4cm}{ImageNet-1K\\(fine-tuning)}}
		& \raggedright V-MoE-S								& 32 & 1 & 81.19 & 96.07 \\
		ImageNet-1K	& \raggedright V-MoGE-S               	& 32 & 1 & \textbf{81.50} & \textbf{96.19} \\
		\cmidrule(lr){2-6}
		(fine-tuning)	& \raggedright SwinV1-MoE-T			& 32 & 1 & 80.81  & 95.86          \\
		& \raggedright SwinV1-MoGE-T                        & 32 & 1 & \textbf{81.06} & \textbf{95.92} \\
		\bottomrule
	\end{tabular*}
\end{table}

\subsection{Image Classification}

\textbf{Setup.} Experiments are conducted on three widely used datasets: CIFAR-100~\cite{krizhevsky2009learning}, Tiny-ImageNet~\cite{le2015tiny}, and ImageNet-1K~\cite{russakovsky2015imagenet}. For CIFAR-100 and Tiny-ImageNet, we apply data augmentation techniques as described in~\cite{seung2021vision}, while adhering to standard practices~\cite{liu2022swin, liu2021swin, wang2021pyramid, liu2022convnet} for ImageNet-1K. We include three popular Transformer models for comparison: Vision Transformer (ViT)~\cite{dosovitskiy2020image}, Swin Transformer V1 (SwinV1)~\cite{liu2021swin}, and Swin Transformer V2 (SwinV2)~\cite{liu2022swin}. 

Consistent with prior studies, the architecture of each expert within the MoE layer is designed to be identical to the feedforward layer it replaces. However, determining the optimal placement of the MoE layer is more of an art than a precise technical process. For the ViT-based MoE models (V-MoE), the MoE layers are positioned in the last two even-numbered blocks, as suggested in~\cite{riquelme2021scaling, videau2024mixture}. In the case of SwinV1-based and SwinV2-based MoE models, the placement strategy varies depending on the dataset. For CIFAR-100 and Tiny-ImageNet, we adopt the approach from~\cite{hwang2023tutel}, placing MoE layers in alternate layers within the last two stages. For ImageNet-1K, following~\cite{videau2024mixture}, MoE layers are inserted exclusively into the last block of each of the last two stages for both SwinV1-based and SwinV2-based MoE models. We conduct evaluations in both training from scratch (i.e., pretraining) and fine-tuning scenarios. For fine-tuning on ImageNet-1K, MoE models perform sparse upcycling to initialize experts from dense checkpoints pretrained on ImageNet-22K~\cite{komatsuzaki2023sparse}. During pretraining, all models are run for 150 epochs, and for fine-tuning, they are trained for 30 epochs, as in~\cite{liu2022swin, liu2021swin}. We incorporate batch prioritized routing and load balancing losses as in~\cite{riquelme2021scaling}. MoGE extends the MoE by introducing our group sparse regularization, while all other aspects remain unchanged. Specifically, MoGE reshapes the representation \(z \in \mathbb{R}^{32}\) into a 2D matrix \(Z \in \mathbb{R}^{4 \times 8}\) and applies a \(3 \times 3\) Gaussian lowpass filter. For more training details, please refer to Appendix~B.

\noindent\textbf{Results.} Table~\ref{tab:ResultsOfImageClassification} presents the top-1 accuracy  (Acc@1) and top-5 accuracy (Acc@5) achieved by various models across different datasets. The larger the value, the better the result, with the best results highlighted in bold. Notably, MoGE consistently outperforms MoE across all datasets in both top-1 and top-5 accuracies. Due to the prohibitive computational cost of pretraining on ImageNet-22K, we instead fine-tune models that are pretrained on ImageNet-22K. We can see that MoGE achieves better top-1 and top-5 accuracy than MoE. The curves of training loss and top-1 accuracy for each model are provided in Appendix~A.

%-------------------------------------------------------------------------
\subsection{Language Modeling}

\textbf{Setup.} Following~\cite{teo2024momentumsmoe}, we evaluate the effectiveness of MoGE on WikiText-103~\cite{merity2016pointer} using three types of MoE models: Switch Transformer (SMoE)~\cite{fedus2022switch}, MomentumSMoE~\cite{teo2024momentumsmoe}, and AdamSMoE~\cite{teo2024momentumsmoe}. MoGE introduces our proposed group sparse regularization to the MoE, while keeping the load balancing loss.

For language modeling, the \mbox{top-$k$} function is typically applied before the softmax function, differing from the sequence introduced in subsection~\ref{sec:VanillaMoE}. Specifically, sparsely activated weights are computed as follows:
	 $$w = \mathrm{softmax}(\mathrm{TopK}(G(x), k)),$$
where the \mbox{top-$k$} function, $\mathrm{TopK}(\cdot,\cdot)$, is defined as:
\begin{align*}
	\mathrm{TopK}(v, k) \!=\!& \begin{cases}v_i, &\text{if } v_i \ \text{is among the top-$k$ elements of } v;\\ -\infty, &\text{otherwise}.\end{cases}.
\end{align*} 
MoGE can  be directly applied to $G(x)$. To facilitate tuning of $\lambda$ in~\eqref{eq:MoGE_finalloss}, we normalize $G(x)$ using the softmax function. Thus, we first compute $z=\mathrm{softmax}(G(x))$ and then pass $z$ into Algorithm~\ref{alg:MoGE} to obtain the group sparse regularization $R(z)$. MoGE reshapes \(z \in \mathbb{R}^{16}\) into a 2D matrix \(Z \in \mathbb{R}^{4 \times 4}\) and applies a \(3 \times 3\) Gaussian lowpass filter. For all MoGE models, we set $\sigma_0=10$, $\sigma_{\min}=1.5$, and $\gamma=0.3$. For further training details, please refer to Appendix~B.

\noindent\textbf{Results.} Table~\ref{tab:ResultsOfWikiText-103} summarizes the perplexity (PPL) values for three types of MoE models on the validation and the test sets of WikiText-103. Lower PPL values indicate better performance, with the best results highlighted in bold. It can be seen that both MoE and MoGE achieve better results with $k=2$ compared to $k=1$. MoGE consistently outperforms MoE across all test sets and most validation sets, clearly demonstrating its superiority. Although MoGE shows slightly inferior performance to MoE on the validation set when using the AdamSMoE-medium model, it still achieves better performance on the more critical test set.  Due to space constraints, the training and the validation PPL curves are provided in Appendix~A.

\begin{table}[t]
	\centering	
	\caption{Comparison of the perplexity (PPL) of MoGE across three types of MoE models.}
	\label{tab:ResultsOfWikiText-103}
	\begin{tabular*}{\linewidth}{l c c c c}
		\toprule
		Model and Type           & \(n\)  & \(k\)  & Valid PPL $\downarrow$ & Test PPL $\downarrow$ \\
		\midrule
		SMoE-small             & 16     & 1      & 90.30             & 91.47                  \\
		SMoGE-small            & 16     & 1      & \textbf{89.08}    & \textbf{89.69}         \\
		\midrule
		SMoE-small             & 16     & 2      & 84.26             & 84.81                  \\
		SMoGE-small            & 16     & 2      & \textbf{82.38}    & \textbf{82.08}         \\
		\midrule
		SMoE-medium            & 16     & 2      & 33.76             & 35.55                  \\
		SMoGE-medium           & 16     & 2      & \textbf{33.50}    & \textbf{35.52}         \\
		
		\midrule\midrule
		MomentumSMoE-small     & 16     & 2      & 85.71              & 86.65      \\
		MomentumSMoGE-small    & 16     & 2      & \textbf{83.10}     & \textbf{83.50}           \\
		\midrule
		MomentumSMoE-medium    & 16     & 2      & 32.29     & 33.46      \\
		MomentumSMoGE-medium   & 16     & 2      & \textbf{31.84}    & \textbf{33.35}            \\
		
		\midrule\midrule
		AdamSMoE-medium        & 16     & 2      & \textbf{31.69 }    & 33.37      \\  
		AdamSMoGE-medium       & 16     & 2      & 31.86     & \textbf{33.28}      \\        
		\bottomrule
	\end{tabular*}
\end{table}

\subsection{Ablation Study}
\label{ablation:StaticVSDynamicSigma}
%For our proposed MoGE, we set the hyperparameter $\lambda$ in Eq.~\eqref{eq:MoGE_finalloss} as described in Appendix~B. %We conduct the following ablation studies for the Gaussian lowpass filter.

\noindent\textbf{Scaling Expert Number $n$.} We analyze the performance of MoGE when scaling $n$. For this investigation, we pretrain the SwinV2-MoE-T and SwinV2-MoGE-T models with 64 or 128 experts on Tiny-ImageNet and ImageNet-1K datasets, as in Table~\ref{tab:TopKandExperts}. The results consistently demonstrate that MoGE outperforms MoE. We also fine-tune SwinV1-MoE-B and SwinV1-MoGE-B with 128 experts on ImageNet-1K dataset, which comprise approximately 1.4B total parameters. As also shown in Table~\ref{tab:TopKandExperts}, MoGE achieves superior performance compared to MoE.

\noindent\textbf{Efficiency.} The additional computation and memory overhead of MoGE relative to MoE mainly depends on the expert number $n$, not the expert size or total model parameters. To validate this, we use four models from Table~\ref{tab:ResultsOfImageClassification} and evaluate them in both pretraining and fine-tuning settings, using $n=32$ and top-1 routing. 
The results are summarized in Table~1 of Appendix~A. The increase in GPU memory usage for MoGE over MoE is negligible. In terms of wall-clock time, pretraining ViT-based MoE models on Tiny-ImageNet with MoGE increases no more than 1\%, while fine-tuning SwinV1-based models on ImageNet-1K increases no more than 1.5\%. Despite large differences in model size (e.g., ViT-S vs. ViT-B or SwinV1-T vs. SwinV1-B), models with the same number of MoE layers exhibit similar absolute increases in training time (e.g., 248 vs. 254 seconds for ViT variants; 211 vs. 226 seconds for SwinV1 variants).

\begin{table}[t]
	\caption{Comparison of MoE and MoGE in scaling $n$.}
	\label{tab:TopKandExperts}
	\centering
	\begin{tabular*}{\columnwidth}{@{\extracolsep{\fill}}c p{2.25cm} cccc@{}}
		\toprule
		Dataset & \centering Model & $n$ & $k$ & Acc@1 & Acc@5 \\
		\midrule
		% ---------------- Tiny-ImageNet ----------------
		\multirow{4}{*}{Tiny-ImageNet}	
		& \raggedright SwinV2-MoE-T							& 128 & 1 & 58.14          & 78.12          \\
		& \raggedright SwinV2-MoGE-T                     	& 128 & 1 & \textbf{59.61} & \textbf{79.09} \\
		\cmidrule(lr){2-6}
		& \raggedright SwinV2-MoE-T							& 128 & 2 & 54.74          & 76.33          \\
		& \raggedright SwinV2-MoGE-T                     	& 128 & 2 & \textbf{59.12} & \textbf{78.91} \\
		\midrule
		
		% ---------------- ImageNet-1K (scratch) ----------------
		%        \multirow{4}{*}{ImageNet-1K}
		\multirow{2}{*}{ImageNet-1K}
		& \raggedright SwinV2-MoE-T							& 64 & 1 & 80.07 & 94.97 \\
		& \raggedright SwinV2-MoGE-T                        & 64 & 1 & \textbf{80.95} & \textbf{95.67} \\
		\midrule
		
		% ---------------- ImageNet-1K (fine-tune) ----------------
		%        \multirow{4}{*}{\parbox[c]{2.4cm}{ImageNet-1K\\(fine-tuning)}}
		ImageNet-1K	& \raggedright SwinV1-MoE-B				& 128 & 1 & 84.88 & 97.50 \\
		(fine-tuning)	& \raggedright SwinV1-MoGE-B		& 128 & 1 & \textbf{85.75} & \textbf{97.98}  \\		
		\bottomrule
	\end{tabular*}
\end{table}

\noindent\textbf{Standard Deviation $\sigma$.} We pretrain the SwinV2-MoGE-T model on CIFAR-100  (see Table~\ref{tab:ResultsOfImageClassification}), with $n=32$ and top-1 routing. With filter size $h=3$, hyperparameters are tuned as follows: we first search for the optimal $\sigma$, then fix $\sigma_0=10$ and set $\sigma_{\min}$ to this value before searching for the best $\gamma$. Using the optimal $\gamma$, we further refine $\sigma_{\min}$. In practice, we use $h=3$, $\sigma_0=10$, and $\gamma=0.3$, with only $\sigma_{\min}$ requiring tuning.

\noindent\textbf{Filter Size $h$.} We use the V-MoGE-S model on Tiny-ImageNet for pretraining. Each MoGE layer uses 128 experts with top-1 routing, yielding a $Z \in \mathbb{R}^{8 \times 16}$ matrix. For both average and Gaussian filters, we set $\lambda=4$e-3 and evaluate different values of $h$. For Gaussian filters, $\sigma_0=10$, $\sigma_{\min}=1.5$, and $\gamma=0.3$ are used. Results are shown in the last two columns of Table~\ref{tab:StaticVSDynamicSigma}, where “3a/5a” and “3g/5g” refer to $3\times3$/$5\times5$ average and Gaussian filters, respectively. Filters with $h=3$ outperform $h=5$, and Gaussian filters consistently surpass average filters.

\begin{table}[h]
	\centering
	\setlength{\tabcolsep}{1.9pt} % 可以微调列间距
	\caption{Ablations on the Gaussian lowpass filter.}
	\label{tab:StaticVSDynamicSigma}
	\begin{tabular}{c c | c c c c | c c c c| c c}
		\toprule
		$\sigma$ & Acc@1 & $\sigma_0$ & $\sigma_{\min}$ & $\gamma$ & Acc@1 & $\sigma_0$ & $\sigma_{\min}$ & $\gamma$ & Acc@1   & $h$ & Acc@1  \\
		\midrule
		3    & 77.25       & 2   & 2  & 0    & 77.57       & 10   & 2     & 0.3   & 77.64           & 3a   & 54.90              \\
		2.5  & 77.50       & 10  & 2  & 3.0  & 77.59       & 10   & 1.5   & 0.3   & \textbf{77.75}  & 3g   & \textbf{55.15}     \\
		2    & \textbf{77.57}       & 10  & 2  & 1.0  & 77.60       & 10   & 1     & 0.3   & 77.72  & 5a   & 54.31 \\
		1.5  & 77.54       & 10  & 2  & 0.3  & \textbf{77.64}       & 10   & 0.5   & 0.3   & 77.23  & 5g   & 54.52          \\     
		\bottomrule
	\end{tabular}
\end{table}

% Conclusion
\section{Conclusion}
\label{sec:conclusion}

This paper introduces a novel perspective on MoE with the widely used \mbox{top-\(k\)} routing, grounded in sparse representation. It proposes applying a group sparse regularization to the routing input, referred to as Mixture of Group Experts (MoGE), indirectly imposes structural constraints on the experts. MoGE further organizes the routing input into a 2D topographic map, grouping neighboring elements to encourage the learning of invariant representations. Consequently, MoGE fosters greater diversity and specialization among experts. Extensive experiments across various image classification and language modeling tasks demonstrate that the proposed MoGE consistently outperforms vanilla MoE, while incurring only minimal increases in computation and memory overhead.

\noindent\textbf{Limitations.}  Our evaluation is currently limited to image classification and language modeling tasks. Designing novel structural constraints in our MoGE framework tailored to specific tasks is a promising direction worth exploring.

\appendix

\section{Additional Results} 
\label{app:results}

\subsection{Efficiency Analysis}
\label{sec:efficiency}

Table~\ref{tab:Efficiency} summarizes the computational and memory comparison between MoE and MoGE. The results show that MoGE introduces a minimal increase in wall-clock time and total GPU memory usage compared to MoE.

\begin{table}[h]
	\centering
	\caption{Comparison of wall-clock time (WCT, in seconds) and total GPU memory cost (TMC, in MB) between MoE and MoGE. }
	\label{tab:Efficiency}
	\begin{tabular*}{\linewidth}{c @{\extracolsep{\fill}}c c c c}
		\toprule
		Dataset & Model & Type & WCT & TMC \\
		\midrule
		\multirow{4}{*}{Tiny-ImageNet}  & \multirow{2}{*}{ViT-S}   & MoE      & ~40617     & ~5295    \\
		~                               & ~                          & MoGE     & ~40865     & ~5299    \\
		\cmidrule(lr){2-5}
		~                               & \multirow{2}{*}{ViT-B}   & MoE      & 112826    & 12071   \\
		~                               & ~                          & MoGE     & 113080    & 12074   \\
		
		\midrule
		~                    & \multirow{2}{*}{SwinV1-T}   & MoE      & ~17343     & ~7163    \\
		ImageNet-1K          & ~                             & MoGE     & ~17554     & ~7167    \\
		\cmidrule(lr){2-5}
		(fine-tuning)        & \multirow{2}{*}{SwinV1-B}   & MoE      & ~30791     & 14547   \\
		~                    & ~                             & MoGE     & ~31017     & 14549   \\
		\bottomrule
	\end{tabular*}
\end{table}

\subsection{Scaling $k$}
\label{sec:scaling_k}

We analyze the performance of MoE and MoGE when scaling $k$ in \mbox{top-$k$} routing. For this investigation, we pretrain the V-MoE-S and V-MoGE-S models with 32 experts on the Tiny-ImageNet dataset. The results are summarized in Table~\ref{tab:TopKandExperts}. We can see that MoGE consistently outperforms MoE. Moreover, MoGE exhibits a less significant performance decline as \(k\) increases. This highlights the ability of MoGE to leverage larger \(k\) values.

\begin{table}
	\centering
	\caption{Comparison of MoE and MoGE across different values of $k$ on image classification.}
	\label{tab:TopKandExperts}
	\setlength{\tabcolsep}{4.5pt}
	\begin{tabular}{c c c c c}
		\toprule
		Type & $k\!=\!1$ & $k\!=\!2$ & $k\!=\!5$ & $k\!=\!9$ \\
		\midrule
		MoE         & 55.38   & 53.46   & 53.58   & 52.07 \\
		MoGE        & \textbf{56.15}   & \textbf{53.75}   & \textbf{54.17}   & \textbf{53.91} \\
		\bottomrule
	\end{tabular}
\end{table}

\subsection{Scheduling Strategy for \(\sigma\)}

\begin{figure}[h]
	\centering
	\includegraphics[width=0.7\linewidth]{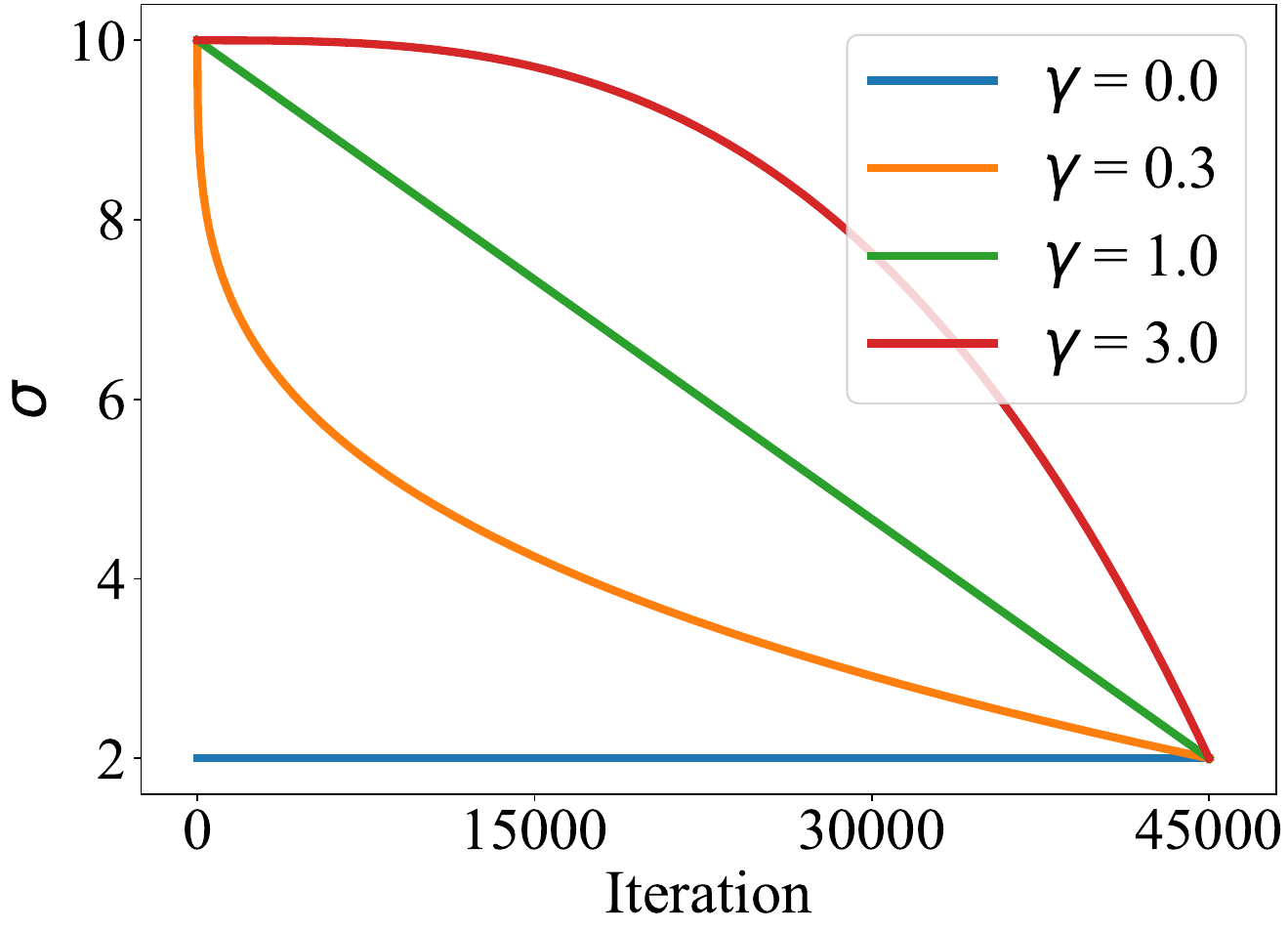}
	\caption{The curves of $\sigma$ with varying $\gamma$ over iterations.}
	\label{fig:DynamicSigma}
\end{figure}

As shown in Eq.~(8), we adjust \(\sigma\) to gradually decay from an initial value \(\sigma_0\) to a minimum value \(\sigma_{\min}\) as the iteration number \(t\) progresses. The rate of this decay is governed by \(\gamma\). Fig.~\ref{fig:DynamicSigma} illustrates the effects of different \(\gamma\) values on the decay behavior of \(\sigma\), with \(\sigma_0=10\), \(\sigma_{\min}=2\), and the total number of iterations \(T=45000\) fixed. Specifically, when \(\gamma=0\), \(\sigma\) remains constant throughout the training process. For \(0 < \gamma < 1\), \(\sigma\) decays rapidly in the early stages and then slows down as training progresses. In the case where \(\gamma = 1\), the decay is strictly linear across all iterations. Finally, for \(\gamma > 1\), the decay starts off slower but accelerates toward the later stages of training.  

Similar to the challenges of designing optimal learning rate scheduling strategies for neural networks, developing an effective $\sigma$ scheduling strategy is equally demanding, as it involves dynamic variations across multiple factors. Based on our empirical studies, it is a good choice to design $\sigma$ scheduling strategies that do not deviate significantly from $\gamma=1$. Future work could draw inspiration from learning rate scheduling techniques in neural networks to further refine $\sigma$ scheduling strategy.

% ======================================================================================================
\subsection{Image Classification}

\begin{figure}[h]
    \centering
    \includegraphics[width=\linewidth]{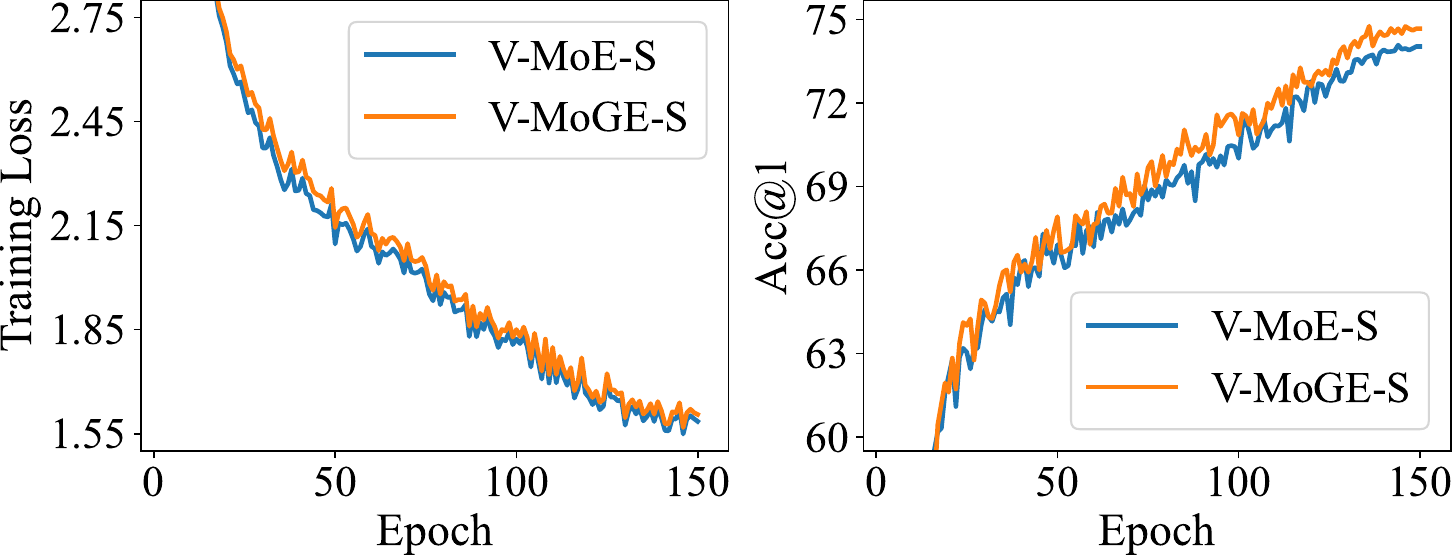}
    \caption{Training loss and top-1 accuracy curves of V-MoE and V-MoGE on CIFAR-100. }
    \label{fig:AppIC_CIFAR100_ViT}
\end{figure}

\begin{figure}[h]
    \centering
    \includegraphics[width=\linewidth]{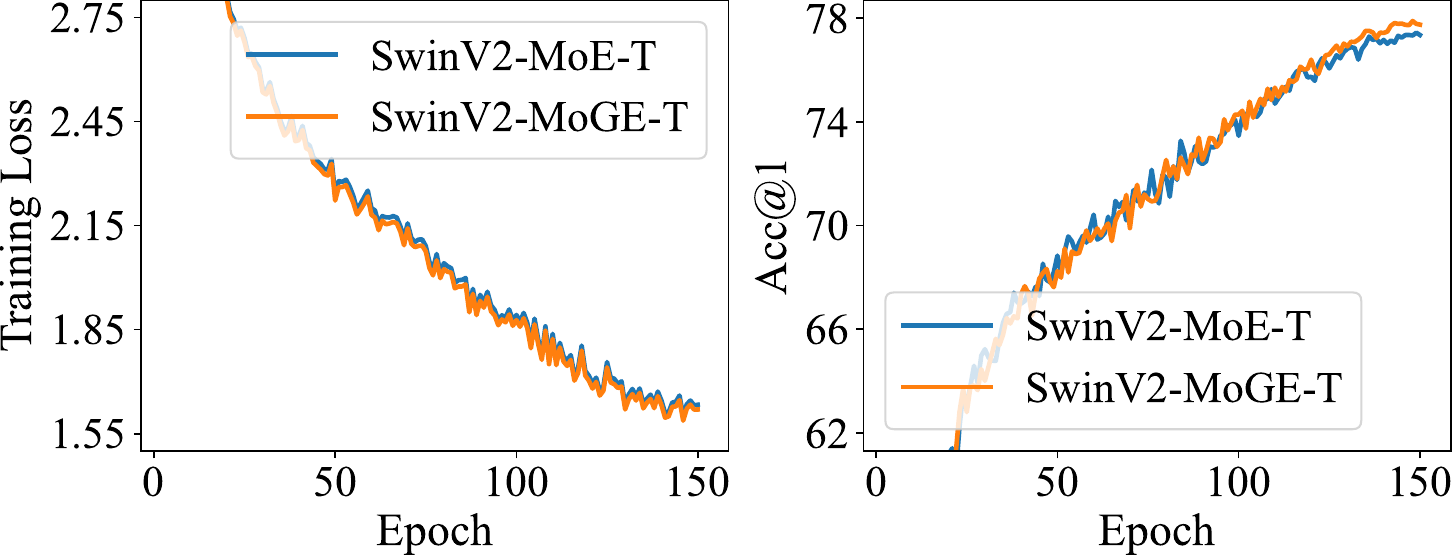}
    \caption{Training loss and top-1 accuracy curves of SwinV2-MoE and SwinV2-MoGE on CIFAR-100.}
    \label{fig:AppIC_CIFAR100_Swin}
\end{figure}

\begin{figure}[h]
	\centering
	\includegraphics[width=\linewidth]{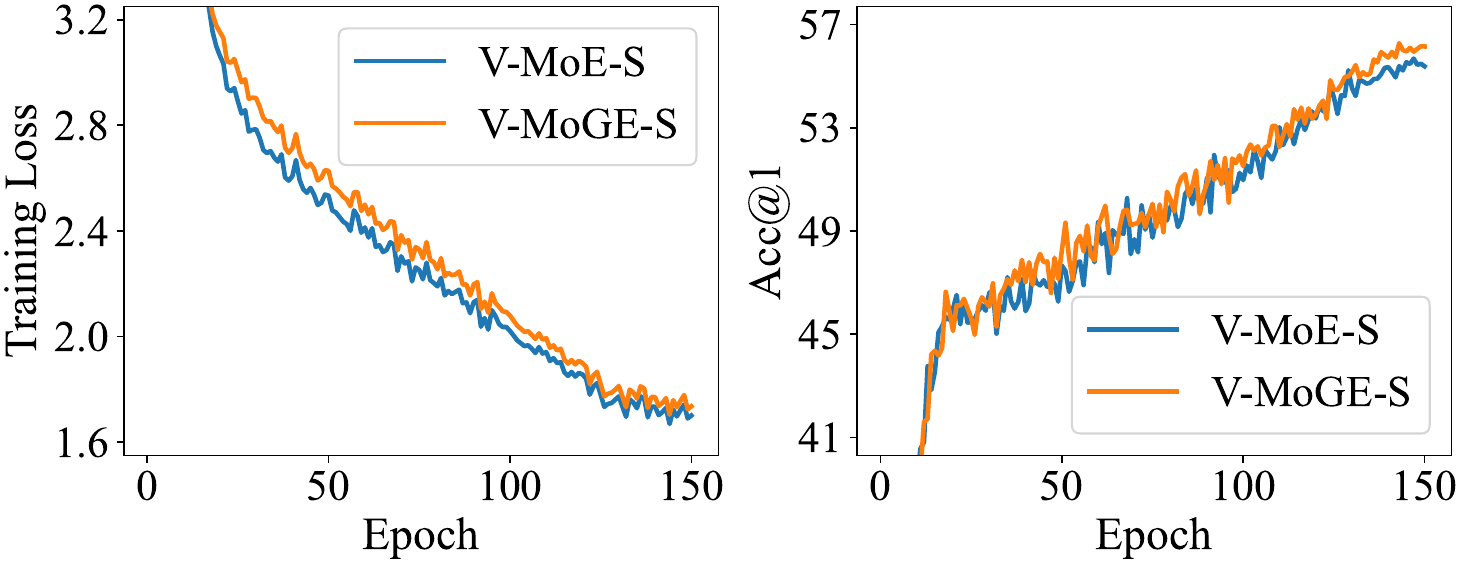}
	\caption{Training loss and top-1 accuracy curves of V-MoE and V-MoGE on Tiny-ImageNet.}
	\label{fig:AppIC_TinyImageNet_ViT}
\end{figure}

\begin{figure}[h]
    \centering
    \includegraphics[width=\linewidth]{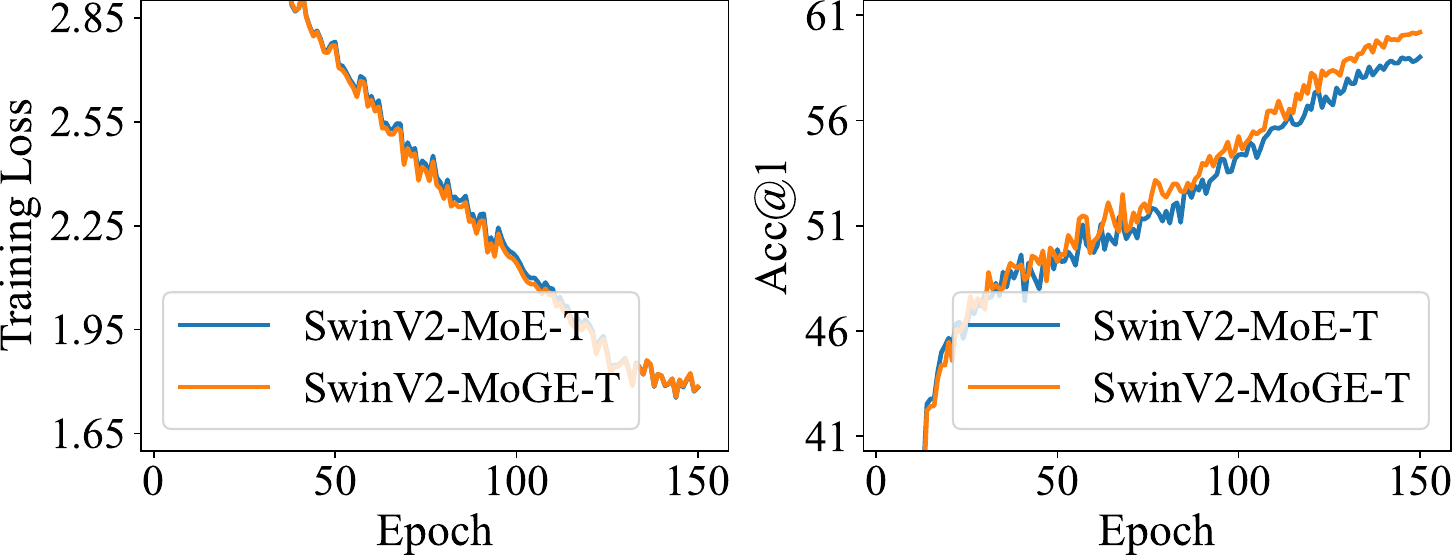}
    \caption{Training loss and top-1 accuracy curves of SwinV2-MoE and SwinV2-MoGE on Tiny-ImageNet.}
    \label{fig:AppIC_TinyImageNet_Swin}
\end{figure}

\begin{figure}[h]
    \centering
    \includegraphics[width=\linewidth]{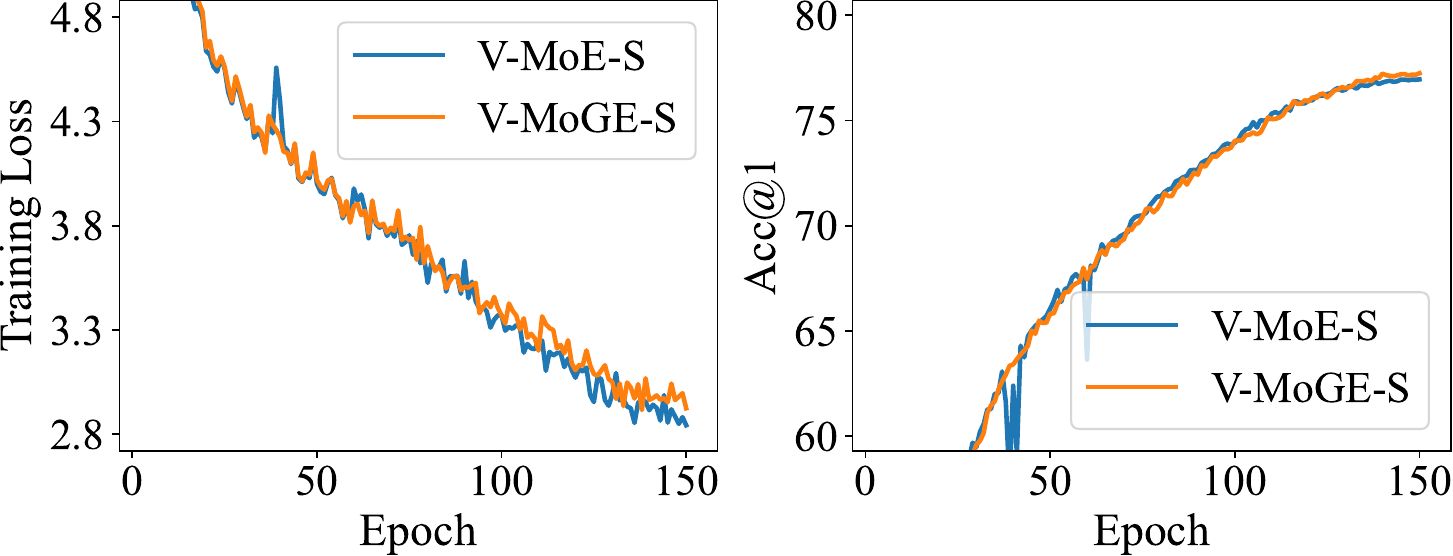}
    \caption{Training loss and top-1 accuracy curves of V-MoE and V-MoGE on ImageNet-1K.}
    \label{fig:AppIC_ImageNet1K_ViT}
\end{figure}

\begin{figure}[h]
    \centering
    \includegraphics[width=\linewidth]{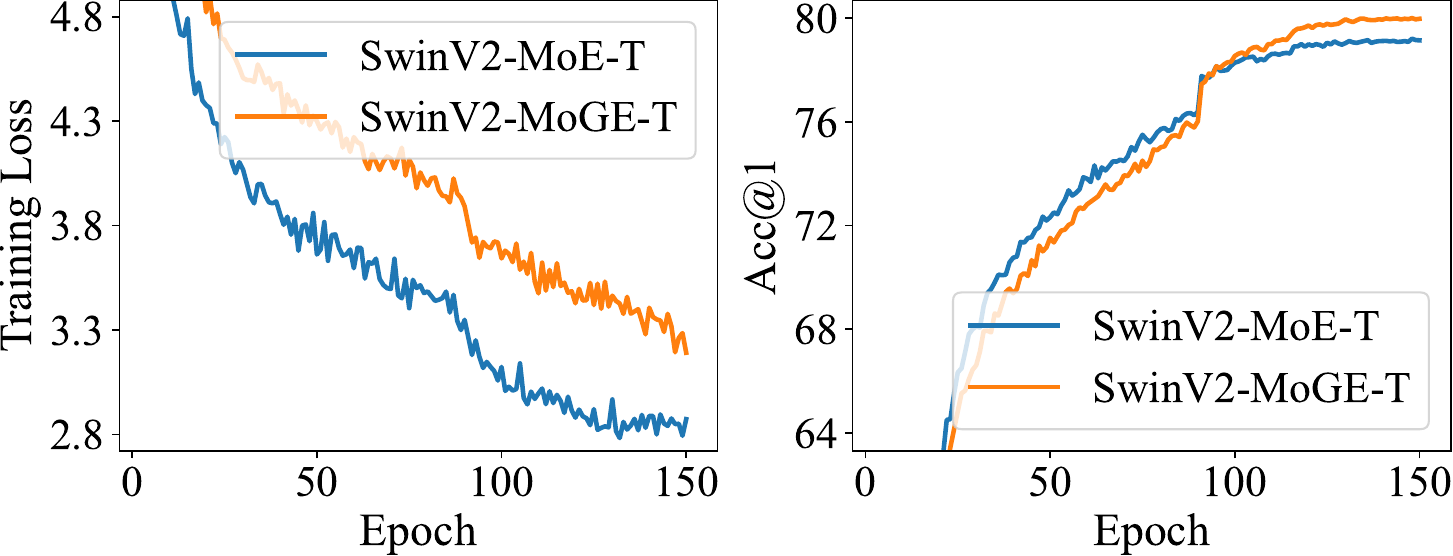}
    \caption{Training loss and top-1 accuracy curves of SwinV2-MoE and SwinV2-MoGE on ImageNet-1K.}
    \label{fig:AppIC_ImageNet1K_Swin}
\end{figure}

\begin{figure}[h]
    \centering
    \includegraphics[width=\linewidth]{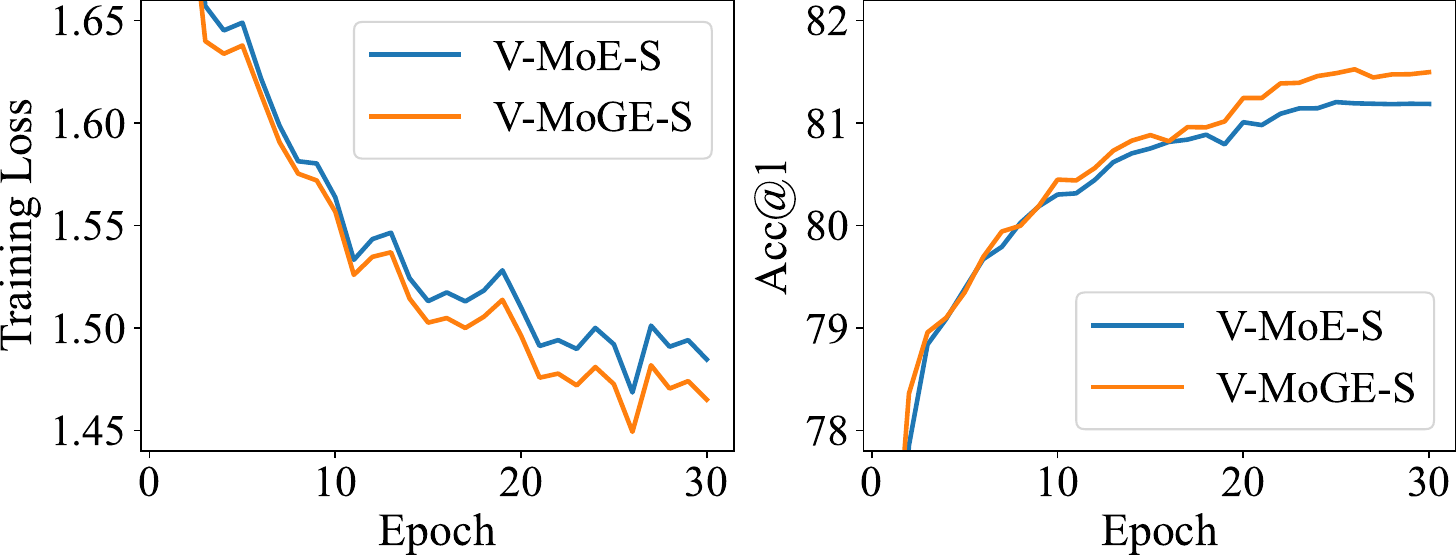}
    \caption{Training loss and top-1 accuracy curves of V-MoE and V-MoGE on ImageNet-1K (fine-tuning).}
    \label{fig:AppIC_ImageNet1KFT_ViT}
\end{figure}

\begin{figure}[h]
	\centering
	\includegraphics[width=\linewidth]{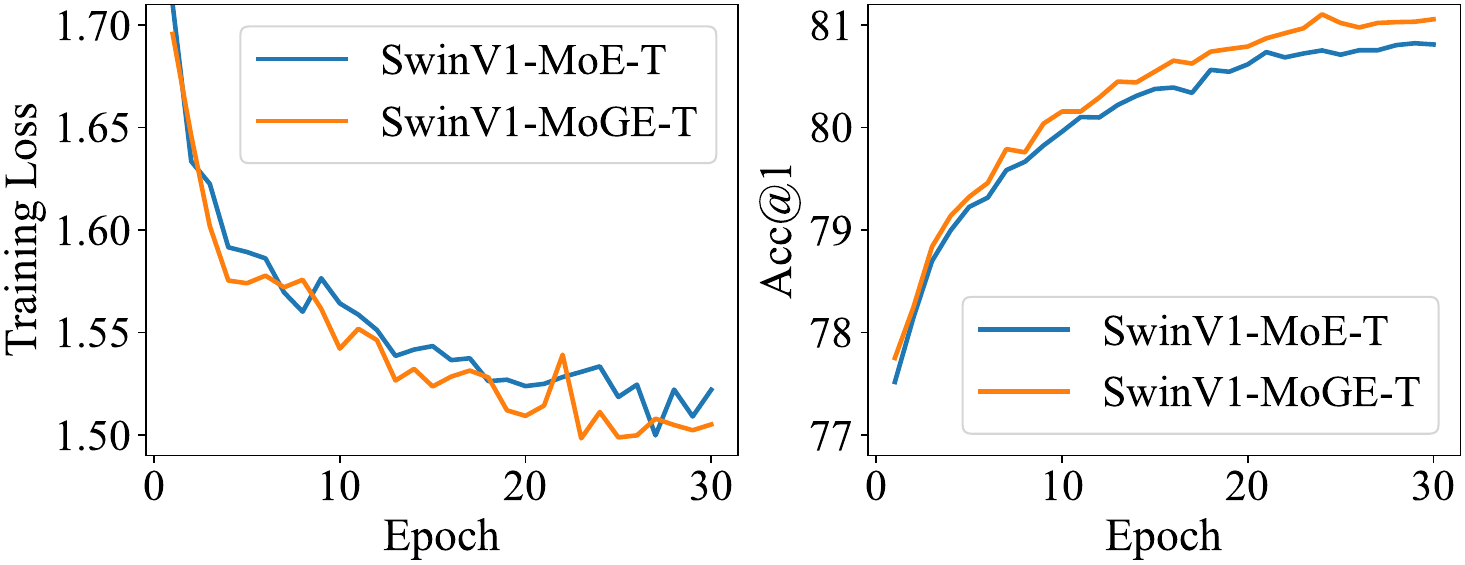}
	\caption{Training loss and top-1 accuracy curves of SwinV1-MoE and SwinV1-MoGE on ImageNet-1K (fine-tuning).}
	\label{fig:AppIC_ImageNet1KFT_Swin}
\end{figure}

Following previous works, we conduct testing using the model trained at the end of training on image classification. The training loss curves and top-1 accuracy curves corresponding to Table~2 of the main paper are presented in Figs.~\ref{fig:AppIC_CIFAR100_ViT}-\ref{fig:AppIC_ImageNet1KFT_Swin}. Consistent with the conclusions in the main paper, MoGE incorporates a regularization constraint compared to MoE. Consequently, its training loss is not always lower than that of MoE. However, MoGE consistently achieves higher top-1 accuracy than MoE.

% ======================================================================================================
\subsection{Language Modeling}

\begin{figure}[h]
    \centering
    \includegraphics[width=\linewidth]{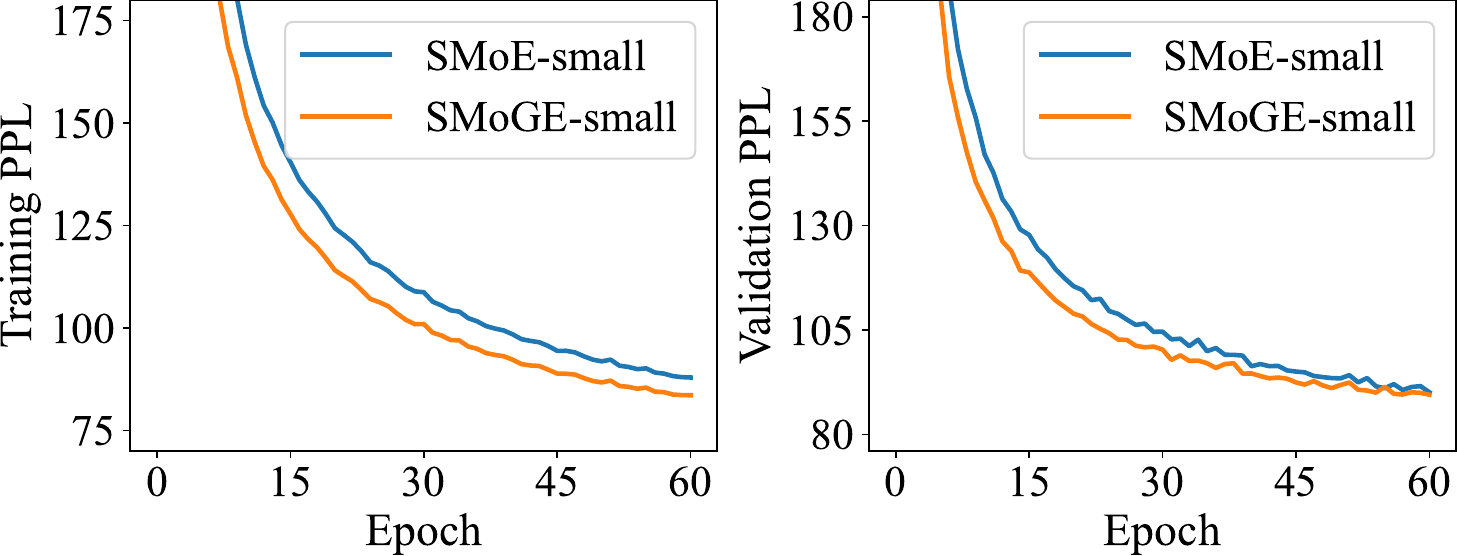}
    \caption{Training and validation PPL curves of SMoE-small and SMoGE-small with $n=16$ and $k=1$.}
    \label{fig:AppLM_WikiText-103_SMoE-small-top1}
\end{figure}

\begin{figure}[h]
	\centering
	\includegraphics[width=\linewidth]{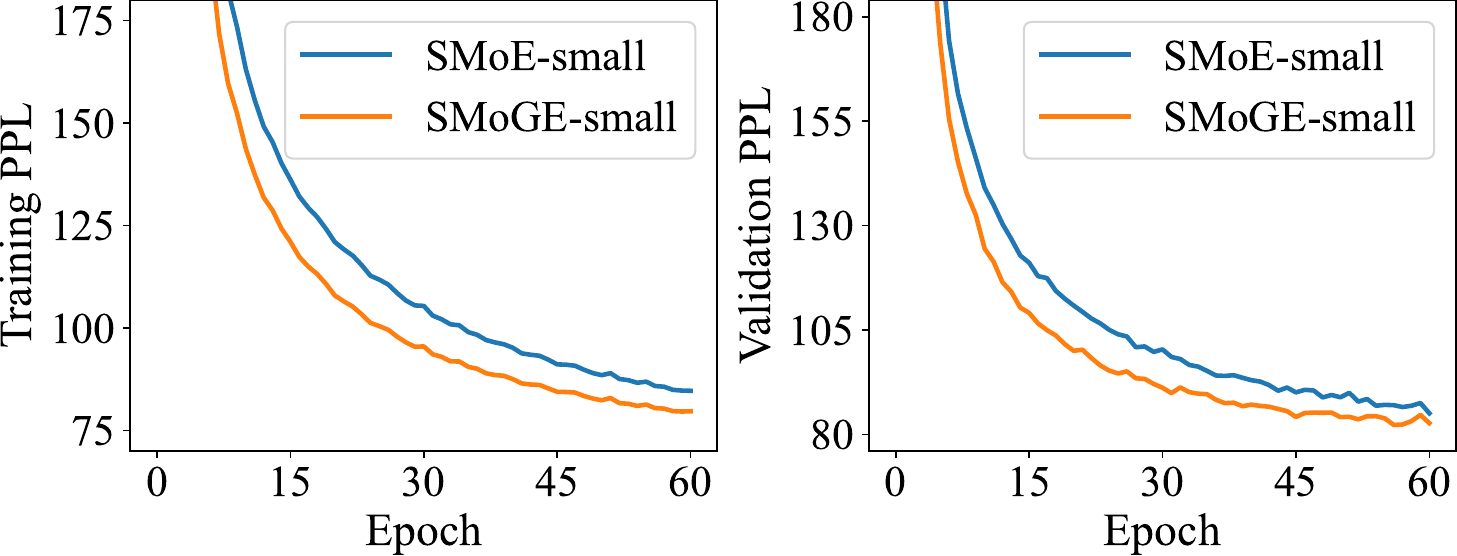}
	\caption{Training and validation PPL curves of SMoE-small and SMoGE-small with $n=16$ and $k=2$.}
	\label{fig:AppLM_WikiText-103_SMoE-small}
\end{figure}

\begin{figure}[h]
    \centering
    \includegraphics[width=\linewidth]{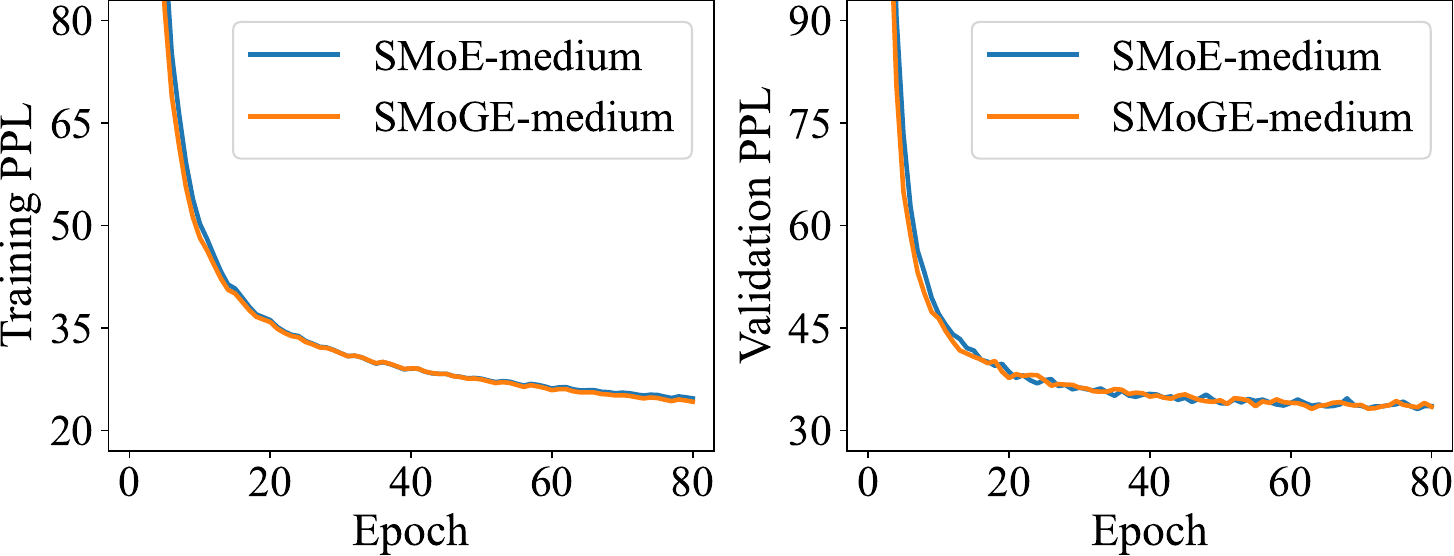}
    \caption{Training and validation PPL curves of SMoE-medium and SMoGE-medium.}
    \label{fig:AppLM_WikiText-103_SMoE-medium}
\end{figure}

\begin{figure}[h]
    \centering
    \includegraphics[width=\linewidth]{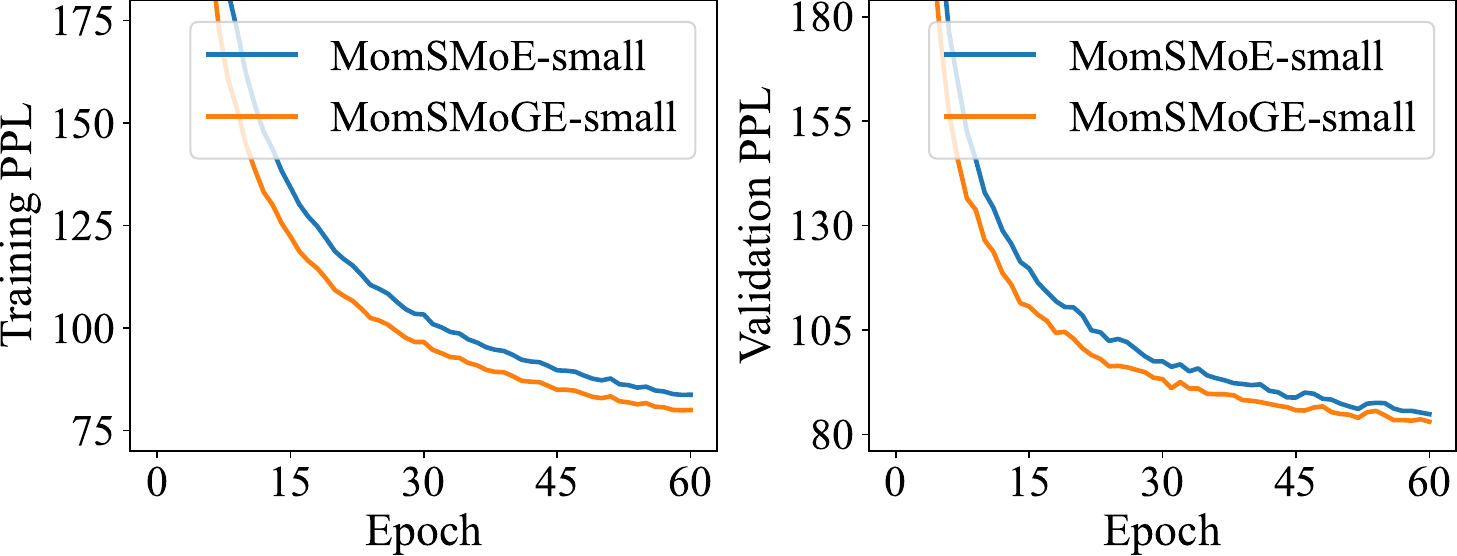}
    \caption{Training and validation PPL curves of MomentumSMoE-small (abbreviated as MomSMoE-small) and MomentumSMoGE-small (abbreviated as MomSMoGE-small).}
    \label{fig:AppLM_WikiText-103_MomentumSMoE-small}
\end{figure}

\begin{figure}[h]
    \centering
    \includegraphics[width=\linewidth]{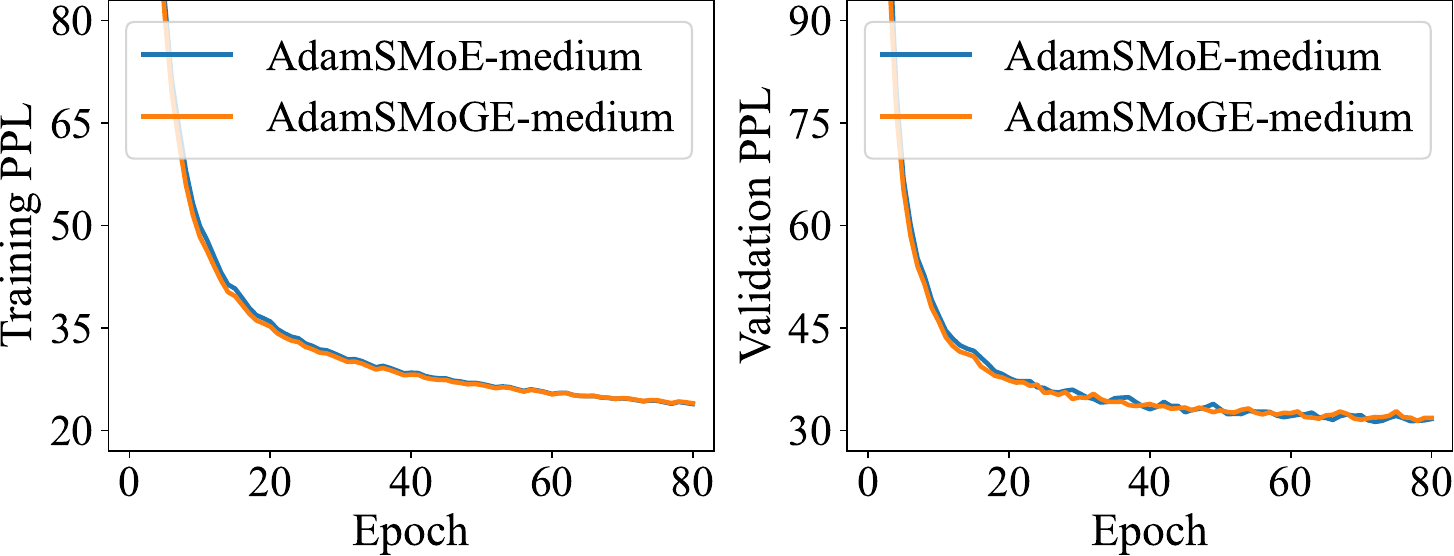}
    \caption{Training and validation PPL curves of AdamSMoE-medium and AdamSMoGE-medium.}
    \label{fig:AppLM_WikiText-103_AdamSMoE-medium}
\end{figure}

Following previous studies~\cite{teo2024momentumsmoe}, we conduct testing using the model that achieves the best performance on the validation set. It is worth noting that performance is evaluated during training only at the end of each epoch rather than after every iteration. Additionally, the test set is evaluated only once, resulting in a single test PPL value. As a result, we do not have test PPL curves. The training and validation perplexity (PPL) curves associated with Table~3 of the main paper are shown in Figs.~\ref{fig:AppLM_WikiText-103_SMoE-small-top1}-\ref{fig:AppLM_WikiText-103_AdamSMoE-medium}. Although MoGE incorporates a regularization constraint compared to MoE, its training PPL remains either superior to or comparable to that of MoE. Likewise, its validation PPL is also either superior to or comparable to that of MoE.

\begin{table*}[t]
	\centering
	\caption{Hyperparameters for image classification.}
	\label{tab:Hyperparameter}
	\begin{tabular}{c|c|c|c|c}
		\toprule
		Settings         & CIFAR-100         & Tiny-ImageNet     & ImageNet-1K          & ImageNet-1K (fine-tuning) \\
		\midrule
		Batch size       & 128               & 128               & 1024                 & 256              \\
		Optimizer        & AdamW             & AdamW             & AdamW                & AdamW            \\
		LR               & 1$\times10^{-3}$  & 1$\times10^{-3}$  & 5$\times10^{-4}$     & 3$\times10^{-5}$ \\
		LR schedule      & cosine            & cosine            & cosine               & cosine           \\
		Weight decay     & 0.05              & 0.05              & 0.05                 & 1$\times10^{-8}$ \\
		Warmup epochs    & 10                & 10                & 15                   & 5                \\
		Epochs           & 150               & 150               & 150                  & 30               \\
		\midrule
		Horizontal flip  & \Checkmark              & \Checkmark              & \XSolidBrush                  & \XSolidBrush              \\
		Random crop      & \Checkmark              & \Checkmark              & \Checkmark                 & \Checkmark             \\
		Auto augment     & \Checkmark              & \Checkmark              & \Checkmark                 & \Checkmark             \\
		Mixup alpha      & 0.25              & 0.25              & 0.8                  & 0.8              \\
		Cutmix alpha     & 0.25              & 0.25              & 1.0                  & 1.0              \\
		Erasing prob.    & 0.25              & 0.25              & 0.25                 & 0.25             \\
		Color jitter     & 0.12              & 0.15              & 0.4                  & 0.4              \\
		\midrule
		Label smoothing  & 0.1               & 0.1               & 0.1                  & 0.1              \\
		Gradient clip    & \XSolidBrush               & \XSolidBrush               & 5.0                  & 5.0              \\
		Loss             & cross-entropy     & cross-entropy     & cross-entropy        & cross-entropy    \\
		\bottomrule
	\end{tabular}
\end{table*}

\begin{table*}
	\centering
	\caption{Training settings of MoGE for image classification.}
	\label{tab:ArchitecturalDetails}
	\begin{tabular}{@{\ }c|cc|cc|cc|cc@{\ }}
		\toprule
		\multirow{2}{*}{Settings} & \multicolumn{2}{c|}{CIFAR-100} & \multicolumn{2}{c|}{Tiny-ImageNet} & \multicolumn{2}{c|}{ImageNet-1K} & \multicolumn{2}{c}{ImageNet-1K (fine-tuning)} \\
		\cline{2-9}
		~                         & V-MoGE      & SwinV2-MoGE      & V-MoGE       & SwinV2-MoGE         & V-MoGE       & SwinV2-MoGE       & V-MoGE         & SwinV1-MoGE \\
		\midrule
		Resolution               & 32          & 32               & 64           & 64                  & 224          & 192               & 224            & 224         \\
		Patch size                & 4           & 2                & 8            & 4                   & 16           & 4                 & 16             & 4           \\     
		Embed                     & 512         & 96               & 512          & 96                  & 512          & 96                & 384            & 96          \\
		Blocks                    & 8           & [2,2,6,2]        & 8            & [2,2,6,2]           & 8            & [2,2,6,2]         & 12             & [2,2,6,2]   \\
		Heads                     & 8           & [3,6,12,24]      & 8            & [3,6,12,24]         & 8            & [3,6,12,24]       & 6              & [3,6,12,24] \\
		Window                    & -           & 4                & -            & 7                   & -            & 12                & -              & 7           \\
		Drop path rate            & 0.1         & 0.1              & 0.1          & 0.1                 & 0.0          & 0.1               & 0.0            & 0.1         \\
		\midrule
		$n$                       & 32          & 32               & 32           & 32                  & 32           & 32                & 32             & 32           \\
		$k$                       & 1           & 1                & 1            & 1                   & 1            & 1                 & 1              & 1           \\
		$\lambda$                 & 4$\times10^{-3}$  & 4$\times10^{-3}$  & 4$\times10^{-3}$  & 4$\times10^{-3}$   & 5$\times10^{-5}$      & 4$\times10^{-3}$   & 4$\times10^{-7}$   & 4$\times10^{-7}$  \\
		$\sigma_0$                & 10          & 10               & 10           & 10                  & 10           & 10                & 10             & 10          \\
		$\sigma_{\min}$           & 1.5         & 1.5              & 1.5          & 1.5                 & 0.5          & 0.5               & 0.5            & 0.5          \\
		$\gamma$                  & 0.3         & 0.3              & 0.3          & 0.3                 & 0.3          & 0.3               & 0.3            & 0.3          \\
		\bottomrule
	\end{tabular}
\end{table*}

\begin{table*}[t]
	\centering
	\caption{Training settings of MoGE for language modeling.}
	\label{tab:ArchitecturalDetailsforLanguageModeling}
	\begin{tabular}{@{\ }c|ccc|ccc@{\ }}
		\toprule
		\multirow{2}{*}{Settings} & \multicolumn{3}{c|}{Small}     & \multicolumn{3}{c}{Medium}                       \\
		\cline{2-7}
		~                         & SMoGE      & SMoGE      & MomentumSMoGE     & SMoGE       & MomentumSMoGE     & AdamSMoGE       \\
		\midrule
		\#Layer                   & 3          & 3          & 3                 & 6           & 6                 & 6               \\    
		Hidden                    & 128        & 128        & 128               & 352         & 352               & 352             \\
		Heads                     & 8          & 8          & 8                 & 8           & 8                 & 8               \\
		Dropout                   & 0.7        & 0.7        & 0.7               & 0.1         & 0.1               & 0.1             \\
		
		\midrule
		Batch size				  & 96		   & 96		   & 96				   & 48			 & 48				 & 48			   \\
		Epochs					  & 60		   & 60		   & 60				   & 80			 & 80				 & 80			   \\
		Optimizer    			  & Adam	   & Adam	   & Adam			   & Adam		 & Adam				 & Adam			   \\
		LR						  & 1$\times10^{-3}$  & 1$\times10^{-3}$  & 1$\times10^{-3}$  & 1$\times10^{-3}$  & 1$\times10^{-3}$  & 1$\times10^{-3}$  \\
		Warmup iterations		  & 3000	   & 3000	   & 3000			   & 4000		 & 4000				 & 4000			   \\
		
		\midrule
		$n$                       & 16          & 16          & 16               & 16           & 16                  & 16           \\
		$k$                       & 1           & 2           & 2                & 2            & 2                   & 2            \\

		$\lambda$                 & 1$\times10^{-6}$  & 1$\times10^{-6}$  & 1$\times10^{-6}$  & 1$\times10^{-6}$  & 1$\times10^{-6}$   & 1$\times10^{-6}$      \\
		$\sigma_0$                & 10          & 10          & 10               & 10           & 10                  & 10           \\
		$\sigma_{\min}$           & 1.5         & 1.5         & 1.5              & 1.5          & 1.5                 & 1.5          \\
		$\gamma$                  & 0.3         & 0.3         & 0.3              & 0.3          & 0.3                 & 0.3          \\
		\bottomrule
	\end{tabular}
\end{table*}

% ======================================================================================================
% ======================================================================================================
\section{Implementation Details}
\label{sup:ImplementationDetails}

\subsection{Datasets}
\label{sup:Datasets}

Because only a single MoE layer is used, we utilize the Fashion-MNIST dataset~\cite{xiao2017fashion} to investigate how MoGE effectively learns invariant representations. For image classification tasks, we employ the CIFAR-100~\cite{krizhevsky2009learning}, Tiny-ImageNet~\cite{le2015tiny}, and ImageNet-1K~\cite{russakovsky2015imagenet} datasets. In language modeling tasks, the experiments are conducted using the WikiText-103 dataset~\cite{merity2016pointer}.

The Fashion-MNIST dataset consists of 70,000 grayscale images, each with a resolution of $28 \times 28$, representing fashion products across 10 categories, with 7,000 images per category. The dataset is divided into a training set of 60,000 images and a test set of 10,000 images.

The CIFAR-100 dataset comprises 60,000 color images, each with a resolution of $32 \times 32$, distributed across 100 classes, with 600 images per class. Among these, 50,000 images are designated for training and 10,000 for testing, resulting in 500 training images and 100 testing images per class.

The Tiny-ImageNet dataset consists of 200 classes, each containing 500 training images, 50 validation images, and 50 test images. All images are resized to $64 \times 64$ pixels. Following~\cite{seung2021vision}, we train our models from scratch using the training set.

The ImageNet-1K dataset comprises 1,000 object classes, including 1,281,167 training images, 50,000 validation images, and 100,000 test images. Following common practices~\cite{liu2022swin, liu2021swin}, most models are trained on the training set using a resolution of $224 \times 224$. Follow~\cite{hwang2023tutel}, a resolution of $192 \times 192$ is adopted for SwinV2-based models~\cite{liu2022swin}.

The WikiText-103 dataset is sourced from Wikipedia articles and is specifically designed to capture long-range contextual dependencies. The training set includes approximately 28,000 articles, amounting to a total of 103 million words, with each article segmented into text blocks of around 3,600 words. The validation and test sets consist of 60 articles each, containing 218,000 and 246,000 words, respectively, for a combined total of approximately 268,000 words.

% ======================================================================================================
\subsection{Training Details for Exploring Invariant Representation Learning}
 
To investigate how MoGE learns invariant representations, we utilize a simplified model comprising a single layer of MoE and MoGE for image classification on the Fashion-MNIST dataset. Both MoE and MoGE models are trained from scratch for 150 epochs, with the parameters for MoGE set as $h=3$, \(\sigma=2\), and \(\lambda = 4\text{e-3}\). 

During training, we use a batch size of 200 and optimize the models using the AdamW optimizer with a learning rate of 0.001. A cosine learning rate scheduler is employed, along with a weight decay of 0.05 and a warmup period of 10 epochs. The training process leverages cross-entropy as the loss function.

\subsection{Training Details for Image Classification}
\label{sup:ImageClassificationHyperparameters}

Tables~\ref{tab:Hyperparameter} and~\ref{tab:ArchitecturalDetails} present detailed information about the experimental setup for our MoGE method in image classification tasks. Specifically, Table~\ref{tab:Hyperparameter} summarizes the hyperparameters used for CIFAR-100, Tiny-ImageNet, ImageNet-1K, and the fine-tuning process on ImageNet-1K. Meanwhile, Table~\ref{tab:ArchitecturalDetails} highlights the training settings for MoGE.

Additionally, we apply the load balancing losses proposed by~\cite{riquelme2021scaling}, with the loss weight consistently set to 0.01. For all experiments, the random seed is fixed at 0. The AdamW optimizer is configured with epsilon = 1e-8 and \(\text{betas} = (0.9, 0.999)\). Furthermore, all model parameters are initialized using a truncated normal distribution within the range \([-2, 2]\), with a mean of \(0\) and a standard deviation of \(0.02\). These configurations are consistently applied to all the compared models.

% ======================================================================================================
\subsection{Training Details for Language Modeling}
\label{sup:LanguageModelingHyperparameters}

For the language modeling tasks, we adopt the settings described in~\cite{teo2024momentumsmoe}. Two model sizes are considered: small (3 layers) and medium (6 layers). The small models are trained for 60 epochs, while the medium models are trained for 80 epochs without incorporating the load balancing loss.  

All model parameters are initialized using the Kaiming uniform distribution~\cite{he2015delving} with \(a=\sqrt{5}\). Additional training details can be found in Table~\ref{tab:ArchitecturalDetailsforLanguageModeling}. Our reproduction of results and implementation of MoGE are based on publicly available source code at: \url{https://github.com/rachtsy/MomentumSMoE}.

% ======================================================================================================
\subsection{Environments}

Our experiments on the Fashion-MNIST, CIFAR-100, and Tiny-ImageNet datasets are conducted using a single NVIDIA A800-SXM GPU with PyTorch 1.13.1+CUDA11.8. For the ImageNet-1K dataset, we utilize four NVIDIA A800-SXM GPUs with PyTorch 1.13.1+CUDA11.8. Similarly, experiments on the WikiText-103 dataset are performed using four NVIDIA A800-SXM GPUs with PyTorch 2.0.1+CUDA11.8.

\bibliographystyle{ACM-Reference-Format}
\bibliography{sample-base}

%\appendix

%Lorem.

\end{document}